\documentclass[runningheads]{llncs}

\usepackage[accept,year=2024,ID=25]{eccv}

\usepackage{eccvabbrv}

\usepackage{graphicx}
\usepackage{booktabs}
\usepackage{multirow}

\usepackage[accsupp]{axessibility}  

\usepackage[pagebackref,breaklinks,colorlinks,citecolor=eccvblue]{hyperref}

\usepackage{orcidlink}

\newcommand\blfootnote[1]{%
\begingroup
\renewcommand\thefootnote{}\footnote{#1}%
\addtocounter{footnote}{-1}%
\endgroup
}

\begin{document}

\title{DAS3D: Dual-modality Anomaly Synthesis \\for 3D Anomaly Detection} 

\titlerunning{Dual-modality Anomaly Synthesis}

\author{Kecen Li \inst{1,3}  \and
Bingquan Dai\inst{4}\and
Jingjing Fu\inst{2,\dagger} \and
Xinwen Hou\inst{1}}

\authorrunning{Kecen Li et al.}

\institute{Institute of Automation, Chinese Academy of Sciences \and
Microsoft Research Asia \and
School of Artificial Intelligence, University of Chinese Academy of Sciences \and Tsinghua Shenzhen International Graduate School}

\maketitle

\begin{abstract}
  Synthesizing anomaly samples has proven to be an effective strategy for self-supervised 2D industrial anomaly detection. However, this approach has been rarely explored in multi-modality anomaly detection, particularly involving 3D and RGB images. In this paper, we propose a novel dual-modality augmentation method for 3D anomaly synthesis, which is simple and capable of mimicking the characteristics of 3D defects. Incorporating with our anomaly synthesis method, we introduce a reconstruction-based discriminative anomaly detection network, in which a dual-modal discriminator is employed to fuse the original and reconstructed embedding of two modalities for anomaly detection. Additionally, we design an augmentation dropout mechanism to enhance the generalizability of the discriminator. Extensive experiments show that our method outperforms the state-of-the-art methods on detection precision and achieves competitive segmentation performance on both MVTec 3D-AD and Eyescandies datasets. The code is available in \url{https://github.com/SunnierLee/DAS3D}.
\blfootnote{ $^{\dagger}$Jingjing Fu is the corresponding author. This work was done when Kecen Li was an intern at Microsoft Research Asia.}
\end{abstract}

\section{Introduction}
\label{sec:intro}

Industrial anomaly detection, which focuses on identifying the anomalous areas of products, plays a pivotal role for inspecting product quality in industrial manufacturing systems. Given the difficulties associated with collecting anomaly data, a significant amount of research efforts have been invested to design unsupervised anomaly detection methods. Most of these methods rely on learning a model of the normal distribution through reconstruction networks~\cite{rudolph2019structuring, kingma2013auto, lecun1989generalization}, or are based on embedding similarity which is extracted from a pretrained network to estimate the normal distribution~\cite{defard2021padim, rudolph2021same, gudovskiy2022cflow,yu2021fastflow}. Recently, knowledge distillation paradigms~\cite{bergmann2019mvtec,deng2022anomaly,Gu_2023_ICCV} are proposed to tackle anomaly detection by flexibly adjusting student's distribution to fit teacher's distribution under the normal samples. A common limitation of these methods is that they exclusively learn the model from anomaly-free data, and are not specifically optimized for discriminative anomaly detection. To resolve this, several attempts~\cite{li2021cutpaste,schluter2022natural} have been made to augment the original anomaly-free samples and transform the unsupervised anomaly detection task into a supervised learning problem. With synthetic anomaly samples, some research~\cite{zavrtanik2021draem, zhang2023destseg} leverages reconstruction networks and supervised-based discriminative networks to attain superior performance in anomaly detection tasks, particularly in terms of accurate localization.

\begin{figure}[t]
    \centering
    \includegraphics[width=0.85\linewidth]{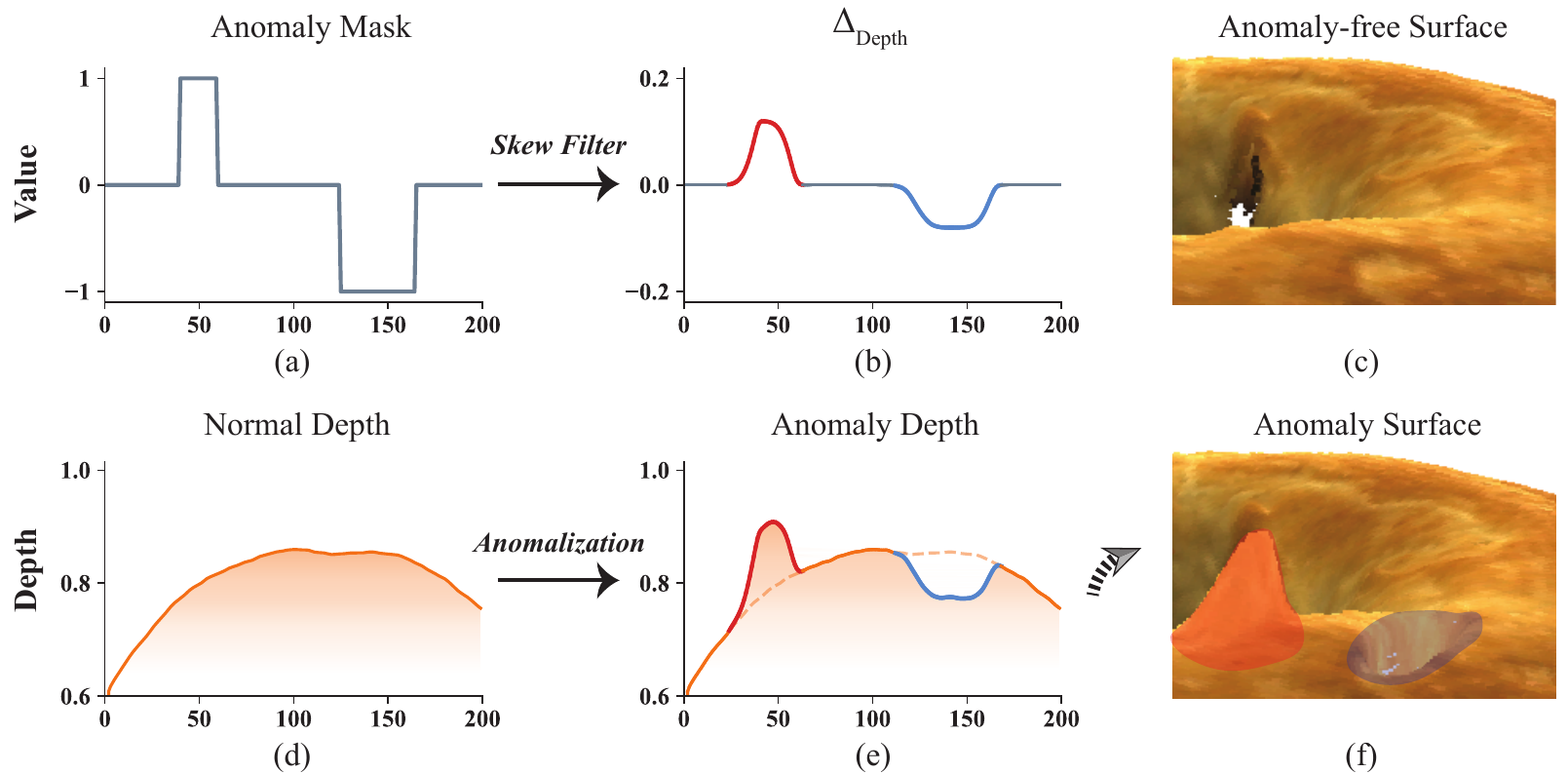}
    \caption{An 1D example of how our 3D anomaly synthesis method works. We first generate an ternary anomaly mask (a), where 1 and -1 indicates the \textcolor[RGB]{236,53,32}{increasing} and \textcolor[RGB]{74,146,206}{decreasing} of depth respectively. We then introduce a skew filter to diversify the shape and magnitude of mask to form the $\Delta_\text{Depth}$ (b). By adding the $\Delta_\text{Depth}$ to the normal depth (d), we generate synthetic anomaly depth (e). We show the change of surface from anomaly-free (c) to the presence of anomalies (f).}
    \label{fig:novelty}
\end{figure}


While synthesizing-based anomaly detection has been comprehensively examined for 2D image data, its potential in 3D anomaly detection is relatively unexplored. This can primarily be attributed to the lack of efficient 3D anomaly synthesis methods. Besides the RGB texture format, point cloud and depth are two frequently used formats for 3D anomaly detection. Directly augmenting 3D data in point cloud format for anomaly synthesis is a complex and expensive task. For instance, in the 3D anomaly detection dataset Eyescandies\cite{bonfiglioli2022eyecandies}, anomalies are synthesized manually via mesh editing. In contrast, EasyNet~\cite{easynet} augments depth by employing a synthetic method of 2D images to create 3D anomalies, but this approach overlooks the physical significance of depth data, leading to unsatisfactory results. To tackle this issue, we introduce a simple and efficient \textbf{D}ual-modality \textbf{A}nomaly \textbf{S}ynthesis for \textbf{3D} anomaly detection (\textbf{DAS3D}), which leverages both depth and RGB format. This method not only considers the spatial characteristics of 3D surfaces, but also augments the dual-modal data simultaneously to create well-aligned anomaly pair in terms of anomalous position. The depth-driven anomaly synthesis process is depicted in Figure~\ref{fig:novelty}. We synthesize anomalies through mathematical manipulations on depth data, thereby economically creating anomalies of various shapes and scales for training. 

Equipped with proposed dual-modal augmentation for anomaly synthesis, we introduce a discriminative anomaly detection network that is trained end-to-end on synthetic dual-modal data. Specially, given a pair of normal RGB and depth images, our dual-modal anomaly generator create anomaly samples that mimic the natural 3D defects. The reconstruction sub-networks reconstruct normal RGB and depth image from the corresponding synthetic anomaly images, separately. Concurrently, the dual-modal discriminator is trained to learns a joint dual-modal reconstruction-anomaly embedding and produces accurate anomaly segmentation maps from the concatenated original and reconstructed features. Note that one of the prevalent challenges in 3D anomaly detection is the effective merging of information from two modalities to enhance detection results. ShapeGuided~\cite{pmlr-v202-chu23b} merely reweights the anomaly maps from two modals for result integration. M3DM~\cite{wang2023multimodal} train two feature transformation networks to transform available dual-modal features into more similar ones for feature fusion. However, the performance of dual-modal results is sometimes inferior to the corresponding single-modal results, suggesting that these dual-modal fusions may not be actively contributing to the final detection process. To fully harness the information from each modal and effectively use it to improve anomaly detection performance, we design an augmentation dropout mechanism to focus on a single modal at times, ensuring a more effective integration of both modals.
Thanks to the design of 3D anomaly synthesis and augmentation dropout mechanism, our method, achieves state-of-the-art performance on MVTec 3D-AD and Eyescandies datasets in terms of image-level AUROC, and also achieves competitive performance in terms of localization accuracy, producing anomaly maps with clearer boundaries and less noises. We summarize the contributions of this work as follows:

\begin{enumerate}
    \item {We propose a multi-modality defect synthesis method capable of economically generating a wide variety of 3D anomalies for 3D synthesizing-based anomaly detection. Our synthesis method is straightforward and effective, and can be integrated with reconstruction-based and distillation-based methods to further delve into 3D anomaly detection.}
    \item {We develop an augmentation dropout mechanism to improve the generalizability of dual-modal discriminator. By randomly introducing single-modal augmentation instances during the end-to-end training process, we further diversify the distribution of the training data and consequently improve anomaly detection performance}
    \item {Our synthesizing-based anomaly detection network is more lightweight compared to existing embedding-based methods, and achieves new state-of-the-art image-level AUROC scores of 0.982 on the MVTec 3D-AD dataset and 0.915 on the Eyescandies dataset.}
\end{enumerate}

\section{Related Work}
Anomaly detection refers to the process of analyzing patterns and features in normal data to identify abnormal data that differs from normal data. Existing anomaly detection methods can be categorised to synthesizing-based method, embedding-based methods, reconstruction-based methods, and distillation-based methods, in terms of their approach to learning normal data patterns and identifying deviations.

\subsection{2D Industrial Anomaly Detection}

\subsubsection{Synthesizing-based Methods}
Synthetic anomaly strategies are frequently utilized in anomaly detection, which augment the original anomaly-free samples to transform the unsupervised anomaly detection task into a supervised learning problem. Prior research efforts \cite{nakazawa2019anomaly, mei2018unsupervised} create anomaly images by introducing random noise pattern to the normal samples, and employ convolutinal denoising autoencoder(AE) network for reconstruction-based anomaly detection.
Contemporary studies have strived to create realistic defective images instead of just using meaningless black-and-white patch images~\cite{li2021cutpaste,schluter2022natural,zavrtanik2021draem,haselmann2019pixel}.
Li et al.~\cite{li2021cutpaste} involve cropping a defect-free region from an original image and pasting it onto a new image at random angles to produce an anomaly image. 
More sophisticated methods utilize background fusion techniques to simulate defects by selecting various background images with different sizes, brightnesses, and shapes. 
For instance, Schlüter et al.~\cite{schluter2022natural} use Poisson fusion, Zavrtanik et al.~\cite{zavrtanik2021draem} select different textured images as defective backgrounds, and Haselmann and Gruber~\cite{haselmann2019pixel} borrow sample synthesis methods in data augmentation. 
Theoretically, the more closely synthetic defects mimic real ones, the more generalizable the image reconstruction and the discriminator become.

\subsubsection{Reconstruction-based Methods}
Reconstruction-based methods are proposed based on an assumption that a model trained on normal data only, cannot represent or reconstruct the anomalies accurately~\cite{zong2018deep,bergmann2018improving}. 
They typically reconstruct samples from the manifold of the training data, using generative adversarial network (GAN)~\cite{lecun1989generalization}, autoencoder (AE)~\cite{rudolph2019structuring}, or variational autoencoder (VAE)~\cite{kingma2013auto}. If the autoencoder generalizes unseen anomaly patterns well, anomalous regions may be reconstructed like normal ones. 
To address this, RIAD~\cite{zavrtanik2021reconstruction} divides the input into disjoint sets, suppressing anomaly generalization. 
SCADN~\cite{yan2021learning} designs inpainting frameworks and trains models on masked normal data to recover unseen regions using context for anomaly detection.
OCRGAN \cite{liang2022omni} decouples images into different frequencies and uses GAN for reconstruction. EdgRec \cite{liu2022reconstruction} achieves good reconstruction results by first synthesizing anomalies and then extracting grayscale edge information from images, which is ultimately input into a reconstruction network. 

\subsubsection{Embedding-based Methods}
Assuming pre-trained networks on large datasets generate distinguishable training features, numerous studies show promising anomaly detection~\cite{deng2009imagenet,schirrmeister2020understanding,Guo_2023_ICCV}. SPADE \cite{cohen2020sub} stores patch-level training features, deriving anomaly scores by measuring distances to test image features. PathCore reduces memory usage through downsampling using greedy coreset subsampling~\cite{roth2022towards}. 
Different from methods using a memory bank, PaDiM~\cite{defard2021padim} abandons slow kNN algorithm and uses Mahalanobis distance metric as an anomaly score. 
There are follow-up methods~\cite{zheng2022focus,huang2022registration} to continue to improve the effectiveness of normal features based on Mahalanobis distance. 

%
\subsubsection{Distillation-based Methods}
Recently, distillation-based method has been well applied in anomaly detection, which identifies anomalies based on the feature differences between the teacher network pretrained on large dataset and the student network that has only seen normal samples. Bergmann \textit{et al.}~\cite{bergmann2019mvtec} first proposed S-T, a distillation paradigm for anomaly detection, which integrates multi-receptive field models. Salehi \textit{et al.}~\cite{salehi2021multiresolution} also detect anomalies by using the difference of multi-level features during distillation. GCAD~\cite{bergmann2022beyond} adopts two pairs of teacher and student to detect both structural anomalies and logical anomalies, which further improves S-T. 
RD4AD~\cite{deng2022anomaly} adopts a reverse distillation framework, which takes the encoder as teacher and the decoder as student, and the student decoder receive teacher encoder to restore teacher feature. 

%
\subsection{3D Industrial Anomaly Detection}

With the swift advancements in industrial inspection devices, the field of 3D anomaly detection has emerged as a relatively new area of research and draw  Bergmann et al. introduce the first public 3D industrial anomaly detection dataset, MVTec 3D-AD~\cite{Bergmann_2022}, which contains both RGB information and point position information for the same manufactured products. Then, Bonfiglioli et al. introduce a synthetic dataset, Eyecandies~\cite{bonfiglioli2022eyecandies}, for unsupervised 3D anomaly detection. Liu et al. introduce Real-3D~\cite{liu2023real3d}, which surpasses MVTec 3D-AD regarding point cloud resolution and 360 degree coverage. 

Given the multitude of methods already proposed in RGB-based anomaly detection, recent research efforts focus on exploring how these established concepts can be adapted and applied to 3D anomaly detection. For Distillation-based methods, Bergmann et al.~\cite{bergmann2023anomaly} introduce a point-cloud feature extraction network of the teacher-student model. 
Rudolph et al.~\cite{RudWeh2023} consider that above method suffers from the similarity of student and teacher architecture and propose asymmetric student-teacher networks (AST).
For reconstruction-based methods, Li et al.~\cite{li2023scalable} use a Mask Reconstruction Network (MRN) to reconstruct the anomaly-free point-cloud whose patches are randomly masked. However, this approach falls short in perfectly reconstructing the anomalous region during the inference stage. For embedding-based method, Wang et al. proposed Multi-3D-Memory (M3DM), where RGB memory, 3D memory and fusion memory are constructed to detect the 3D defects. They use the predominate feature extractors, ViT and PointMAE to extract the RGB and 3D features respectively and obtain fusion memory through additionally training two feature transformation networks.

There have been few synthesizing-based methods, despite their notable success in RGB-based anomaly detection tasks. Recently, Chen et al.~\cite{easynet} and Zavrtanik et al.~\cite{CD} propose novel depth anomaly generation methods. However, their methods do not consider the properties of real 3D defects, thereby resulting in subpar prediction and localization precision. In this paper, we propose a novel 3D anomaly generation method based on depth image. Furthermore, we introduce an augmentation dropout mechanism to encourage our discriminator to deliver more precise predictions.

\begin{figure}[t]
    \centering
    \includegraphics[width=0.9\linewidth]{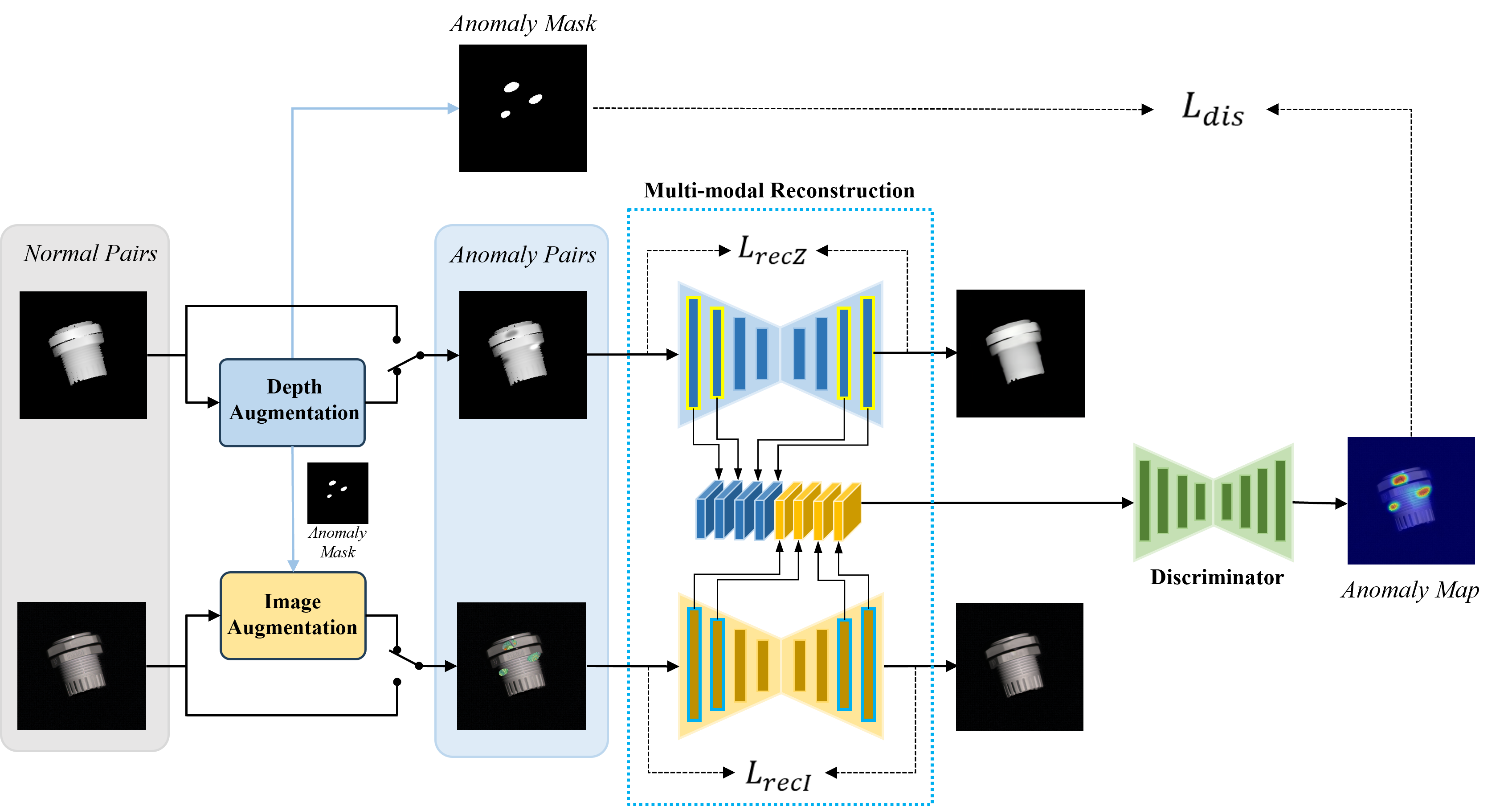}
    \caption{The framework of our method. Given a paired normal depth and RGB images, we design a novel dual-modal anomaly synthesis method to generate the anomaly sample and the corresponding anomaly mask. We train two reconstruction networks to restore the anomaly depth and RGB images to their normal ones. The features in shallow layers from two reconstruction networks are concatenated as the input of an anomaly discriminator, which is trained to predict the anomaly mask. To enhance generalizability of the discriminator, we design an augmentation dropout mechanism to randomly set one anomaly modal into its normal one.}
    \label{fig:framework}
    \vspace{-10.0pt}
\end{figure}

\section{Methods}

In this section, we first detail the overview of DAS3D in Section \ref{sec3.2}. Next, in Section \ref{sec3.3}, we elaborate on our proposed anomaly synthesis pipeline and the implementation details for depth and RGB images. Finally, in Section \ref{sec3.4}, we introduce the dual-modal discriminator. Before next Section, we first give a problem definition. Given a set of anomaly-free training examples $\mathcal{T} = \{\left(I_i, Z_i\right)\}_{i=1}^{N}$, where the $I_i$ and $Z_i$ are the $i$-th paired RGB and depth image, respectively. Our objective is to develop a dual-modal anomaly detector from these samples. When a normal or abnormal sample is presented during the testing phase, the detector should be capable of determining whether the object contains anomalies and identifying the location of the anomaly, if any anomaly is detected. 




\subsection{Dual-modal Anomaly Detection Framework}
\label{sec3.2}

As illustrated in Figure~\ref{fig:framework}, the proposed method is composed of a dual-modal anomaly generator, two reconstruction sub-networks, $F_{I}$ and $F_Z$ for RGB and depth image reconstruction respectively, and a dual-modal discriminator $D$. Given a paired normal RGB and depth images, our dual-modal anomaly generator can generate anomaly samples, which effectively mimic the natural 3D defects and can be processed efficiently. In order to enhance the diversity of synthetic pairs, an augmentation drop-out module is employed to randomly drop the augmentation. Using these synthetic anomaly samples, the reconstructive sub-networks are trained separately to implicitly detect these synthetic anomalies through reconstructing the normal RGB and depth images respectively. Simultaneously, the dual-modal discriminator learns a joint dual-modal reconstruction-anomaly embedding and produces accurate anomaly segmentation maps from the concatenated dual-modal reconstruction features. 

\subsection{Dual-modal Augmentation}
\label{sec3.3}
\begin{figure}[t]
    \centering
    \includegraphics[width=0.9\linewidth]{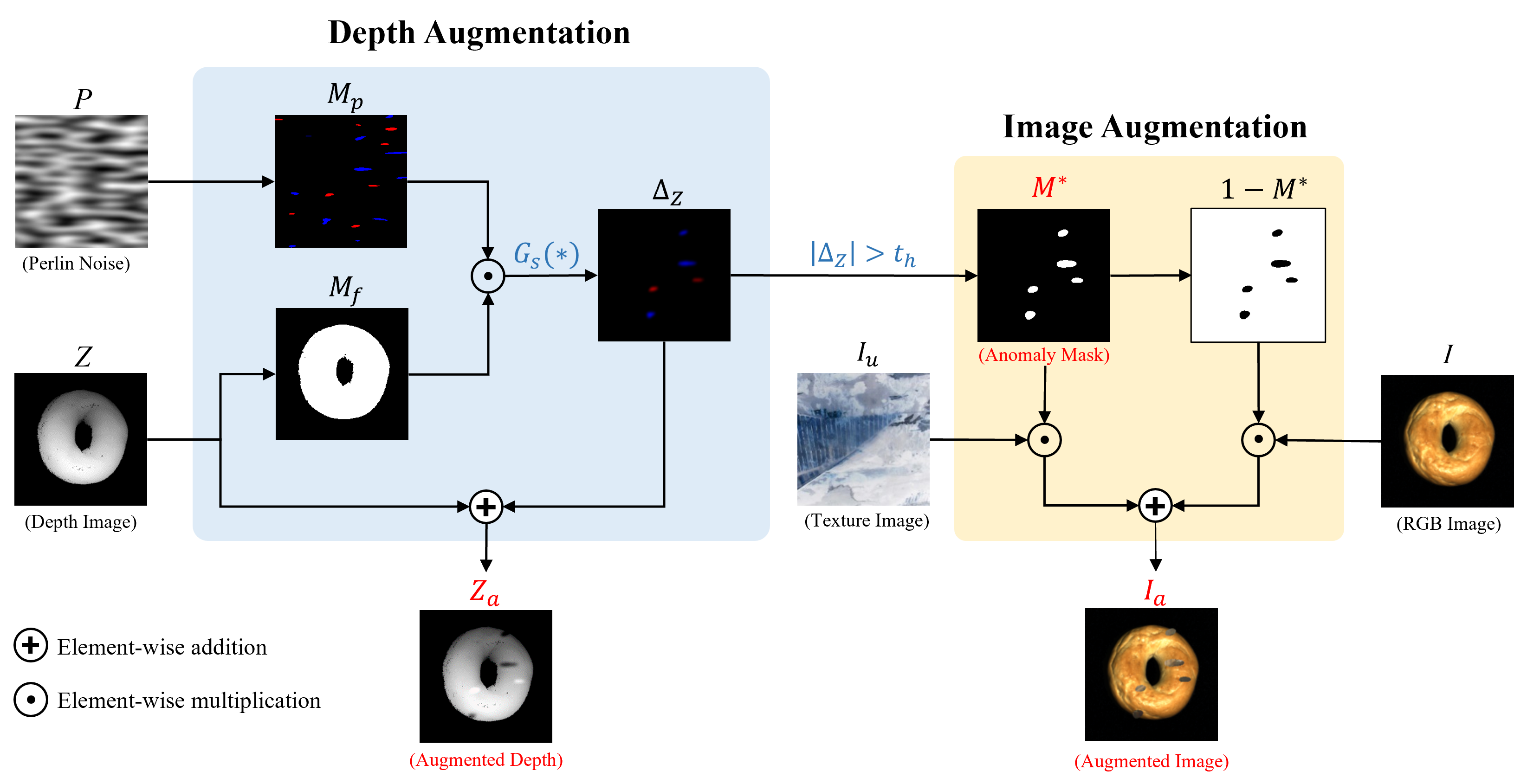}
    \caption{The pipline of our dual-modal augmentation method. We first sample a Perlin noise image $P$ to obtain a ternary mask $M_p$. We obtain a foreground mask $M_f$ from the depth image to remove the background region of $M_p$. We introduce a skew Gaussian filter $G_s$ to operate the foreground-only defects map to obtain the depth change $\Delta_Z$, which is added to the depth image to obtain our augmented depth image. The $\Delta_Z$ can be binarized with a threshold $t_h$ to obtain our anomaly mask $M^*$. The augmented RGB image is generated through randomly mixing the anomaly-free RGB image and a texture image.}
    \label{fig:generator}
    \vspace{-10pt}
\end{figure}

Given a paired normal RGB and depth images $\left(I,Z \right)$, our aim is to generate the anomaly samples $\left(I_a,Z_a \right)$, which attempts to emulate the natural 3D defects. Depth anomaly generation, unlike RGB anomaly generation, necessitates thoughtful deliberation about the nature of 3D defects and the consistency between 3D and RGB. When augmenting the depth data, we primarily focus on three aspects: a) The defects in 3D data typically appear as concave and convex surfaces, which correspond to an increase and decrease in depth values respectively. b) From a physical perspective,  most anomalous surfaces exhibit continuity. c) The diversity of anomalous surfaces influences the amplitude and curvature of the convex and concave surfaces.

\noindent \textbf{Depth Augmentation.}
As illustrated in Figure~\ref{fig:generator}, we first extract the foreground mask $M_f$ from the depth image with a threshold $t_f$, which can be estimated through averaging the depth of background region. Following a RGB-based random mask generation method~\cite{zavrtanik2021draem}, we generate a noise image $P$ using a Perlin noise generator~\cite{perlin}, and then ternarize this noise image with a threshold $t_p$ to acquire a ternary mask. 
\begin{equation}
\label{eq:Md}
    M_p\left[i,j\right] = \begin{cases}
        -1, &  P \left[i,j\right]<-t_p\\
        1, &  \;\;P \left[i,j\right]>t_p\\
        0, & \;\;\;\;\;\; \text{others}\\
    \end{cases}.
\end{equation}

\noindent The values in ternary mask $M_p$ actually indicate the concave and convex surfaces in the sequential augmentation process, where regions with value of -1 is for concave augmentation and colored in blue, while the regions with value of 1 is for convex and colored in red, as shown in Figure~\ref{fig:generator}. Ternary anomaly mask $M_p$ is multiplied by foreground mask $M_f$ to extract the augmentation mask within foreground regions, denoted as $M_t$. 
\begin{equation}
\label{eq:mt}
    M_t = M_f \odot M_p,
\end{equation}
where $\odot$ represents element-wise matrix multiplication. To emulate real object defects, often presenting as convex or concave surfaces,
we introduce a skew Gaussian filter to smooth the edge of the mask $M_t$. To formalize this filter, we first give the probability density function (PDF) of 2-D skew normal distribution from~\cite{skewnormal} as

\begin{equation}
    f_{s}\left(\mathbf{x};\alpha,\Sigma\right) = 2 \phi_{2} \left(\mathbf{x};0,\Sigma \right) \Phi \left(\alpha^\text{T}\mathbf{x}\right), \;\;\; \mathbf{x} \in \mathbb{R}^2,
\end{equation}

\begin{figure}[t]
    \centering
    \includegraphics[width=1\linewidth]{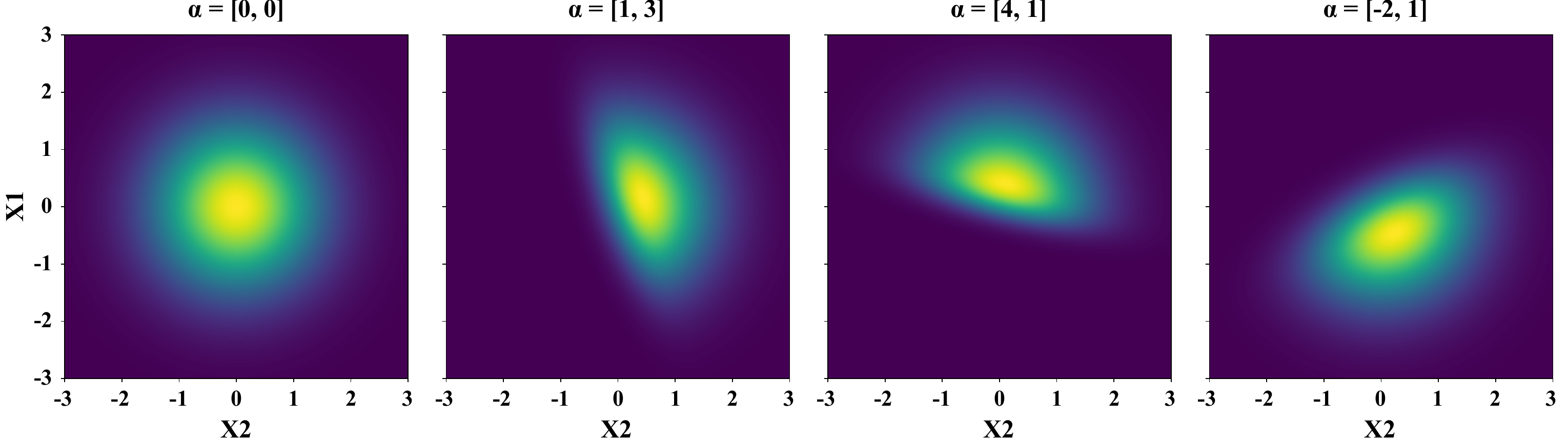}
    \caption{The PDF of 2D skew normal distribution with different $\alpha$ and $\Sigma=I$. The peak of PDF is changed with various $\alpha$, which diversifies our shape of synthetic defects.}
    \label{fig:skew}
    \vspace{-10pt}
\end{figure}

\noindent where $\phi_{2} \left(\mathbf{x}; 0,\Sigma \right)$ signifies the two-variate normal density with zero mean and correlation matrix $\Sigma$. Here, $\Phi \left(\alpha^\text{T}\mathbf{x}\right)$ is the cumulative distribution function (CDF) of the univariate Gaussian and $\alpha$ is the skew coefficient. 
Our skew Gaussian filter is defined as

\begin{equation}
    \label{eq:skewfilter}
    G_{s,h\times h}\left[i,j\right] = f_{s}\left(\left (i-h/2,j-h/2\right);\alpha,\Sigma\right),
\end{equation}
where $h$ is the kernel size of the skew Gaussian filter, which is determined by $\Sigma$. The skew coefficient $\alpha$ is randomly sampled from the uniform distribution $U\left(-0.5,0.5\right)$. The final defect $\Delta_Z$ is defined as

\begin{equation}
    \label{eq:conv}
    \Delta_Z = \text{norm}\left(G_{s,h \times h}\right) \ast M_t,
\end{equation}
where $\text{norm}(\cdot)$ and $\ast$ represent normalization and convolution operation, respectively. The final synthetic anomalous depth image $Z_a$ is defined as 

\begin{equation}
    Z_a = Z + p_z\Delta_Z,
\end{equation}
where $p_z$ is sampled from $U\left(p_\text{min},p_\text{max}\right)$, indicating the range of defects magnitude. Instead of replacing the original depth with the synthetic defect, we treat the synthetic defect as a change in depth and add it to the original. This approach implicitly consider the real geometric properties of different objects on different surfaces and enhance the diversity of our synthetic defects. Due to the convolution operation from Equation~\ref{eq:conv}, the actual region of depth change does not align with the original anomaly mask $M_t$ from Equation~\ref{eq:mt}. As a result, we refine the anomaly mask as

\begin{equation}
\label{eq:refinedMask}
    M^* \left[i,j\right] = \begin{cases} 
        0, & p_z|\Delta_Z\left[i,j\right]|<t_h\\
        1, & \;\;\;\;\;\;\;\; \text{others}\\
    \end{cases}.
\end{equation}

\noindent \textbf{Image Augmentation.}
For RGB defect generation, we expect the defects on RGB images can be consistent with those on depth images. Therefore, we use the refined anomaly mask $M^*$ from Equation~\ref{eq:refinedMask} to determine the defect region of RGB image. To modify the corresponding pixel region of normal RGB image, we first sample a texture image $I_u$ from a pre-constructed image dataset \cite{DTD} which is unrelated to our training set. We can generate a naive anomaly RGB image as $(1-M^*) \odot I + M^* \odot I_u$. However, this change is usually too sharp. We generate a random mix coefficient $\beta$ from $U\left(0,0.8\right)$, which can be used to smooth the change. The final augmented RGB image is defined as 

\begin{equation}
\label{RGBAug}
    I_a = \left(1-M^* \right) \odot I + M^* \odot \left( \left(1-\beta \right) I_u + \beta I\right).
\end{equation}

\subsection{Dual-modal Discriminator}
\label{sec3.4}
With synthetic anomaly samples, we need to train an anomaly discriminator to perform anomaly score prediction and anomaly localization. We first introduce two reconstruction modules, $F_I$ and $F_Z$, which are two UNet-like networks. We train these two networks to reconstruct the anomalous regions towards normal while preserving other normal regions of the RGB and depth image respectively. For depth reconstruction, we employ classical $L2$ loss to minimize the discrepancy between the normal depth image $Z$ and reconstructed depth images $F_{Z}\left(Z_a\right)$ as:

\begin{equation}
    \label{eq:recZloss}
    L_{recZ}\left(Z,Z_a\right)= \Vert Z-F_Z\left(Z_a\right)\Vert_2^2.
\end{equation}
For RGB reconstruction, in additional to $L2$ loss, we also use the SSIM loss to minimize the perceptual
differences.
Thus, our loss function for RGB reconstruction is defined as

\begin{equation}
    \label{eq:recIloss}
    L_{recI}\left(I,I_a\right)= \Vert I-F_I\left(I_a\right)\Vert_2^2 + L_\text{SSIM}\left(I,F_I\left(I_a\right)\right).
\end{equation}

With two reconstruction networks, we can extract the features of RGB and depth images respectively for training our anomaly discriminator. 
Specifically, we choose the first two layer features and last two layer features. These features are upsampled to the same resolution and then concatenated as our fused feature, denoted as $f_\text{fuse}$. Our discriminator is also a UNet-like network, which is trained to predict an accurate anomaly map. Since the region of defect is typically small, we use Focal loss~\cite{lin2017focal} to reduce the discrepancy between the prediction of our discriminator and the ground-truth as

\begin{equation}
    \label{eq:predloss}
    L_{dis}\left(f_\text{fuse},M^*\right) = L_\text{Focal}\left(D\left(f_\text{fuse}\right),M^*\right).
\end{equation}

\noindent \textbf{Augmentation Dropout Mechanism.} We jointly train the reconstruction networks and anomaly discriminator using the three loss functions mentioned above. Our 3D anomaly generator products a paired samples, where both RGB and depth images exhibit abnormalities. However, in real-world scenarios, we frequently encounter situations where an object has defects, but its depth or RGB image appears normal. For instance, a potato may have stains that are not typically visible in its depth image. To create more diverse anomaly data and enhance the generalizability of the dual-modal discriminator, we design an augmentation dropout mechanism to enforce the anomaly discriminator to handle such cases. Specifically, we sample two drop signal $d_I$ and $d_Z$ from $B\left(1,p_d\right)$, which represents a Bernoulli distribution with drop probability as $p_d$. Consequently, our anomaly samples and anomaly mask can be redefined as

\begin{equation}
\label{eq:dropout}
    \Tilde{I}_a = d_{I}I+\left(1-d_{I}\right)I_a,\Tilde{Z}_a = d_{Z}Z+\left(1-d_{Z}\right)Z_a,\Tilde{M}^* = \left(1-d_{I} d_{Z}\right) M^*.
\end{equation}
The above equation implies that an object is considered normal only when both modals (RGB and depth) are normal. Our final loss function is defined as

\begin{equation}
    \label{eq:lossfunction}
    L=L_{recZ}\left(Z,\Tilde{Z}_a\right)+L_{recI}\left(I,\Tilde{I}_a\right)+L_{dis}\left(f_\text{fuse},\Tilde{M}^*\right).
\end{equation}

\begin{table}[t]
\fontsize{6}{8}\selectfont
\setlength{\tabcolsep}{0.6mm}
\renewcommand{\arraystretch}{1.1}
    \centering
    \caption{I-AUROC score for anomaly detection of MVTec 3D-AD. The best is in red.}
    \label{tab:i-auroc-mvtec}
    \begin{tabular}{cl|cccccccccc|c}
    \toprule
    \multicolumn{2}{c|}{Method} & Bagel & Cable & Carrot & Cookie &Dowel & Foam & Peach & Potato & Rope & Tire & Mean\\[1pt]    \hline 
    \multirow{6}{*}{RGB only} & 
     BTF~\cite{horwitz2022feature} & 0.854 & 0.840 & 0.824 & 0.687 & 0.974 & 0.716 & 0.713 & 0.593 & 0.920 & 0.724 & 0.785\\
     &AST\cite{RudWeh2023} & 0.947 & 0.928 & 0.851 & 0.825 & 0.981 & \textcolor{red}{0.951} & 0.895 & 0.613 & \textcolor{red}{0.992} & 0.821 & 0.880\\
     & EasyNet~\cite{easynet} & \textcolor{red}{0.982} & \textcolor{red}{0.992} & 0.917 & 0.953 & 0.919 & 0.923 & 0.840 & \textcolor{red}{0.785} & 0.986 & 0.742 & 0.904\\
     & M3DM~\cite{wang2023multimodal}& 0.944 & 0.918 & 0.896 & 0.749 & 0.959 & 0.767 & 0.919 & 0.648 & 0.938 & 0.767 & 0.850 \\
     & ShapeGuided~\cite{pmlr-v202-chu23b} & 0.911 & 0.936 & 0.883 & 0.662 & 0.974 & 0.772 & 0.785 & 0.641 & 0.884 & 0.706 & 0.815\\
     & 3DSR~\cite{CD} & 0.844 & 0.930 & \textcolor{red}{0.964} & 0.794 & \textcolor{red}{0.998} & 0.904 & \textcolor{red}{0.938} & 0.730 & 0.978 & \textcolor{red}{0.900} & 0.898\\
     & DAS3D & 0.981 & 0.974 & 0.916 & \textcolor{red}{0.960} & 0.959 & 0.950 & 0.888 & 0.750 & 0.982 & 0.817 & \textcolor{red}{0.918}\\ [2pt]
     \hline
    \multirow{6}{*}{3D only} & BTF\cite{horwitz2022feature} & 0.696 & 0.553 & 0.824 & 0.696 & 0.795 & 0.773 & 0.573 & 0.746 & 0.936 & 0.553 & 0.714\\
     & AST\cite{RudWeh2023} & 0.881 & 0.576 & 0.965 & 0.957 & 0.679 & 0.797 & 0.990 & 0.915 & 0.956 & 0.611 & 0.833 \\
     & EasyNet~\cite{easynet} & 0.735 & 0.678 & 0.747 & 0.864 & 0.719 & 0.716 & 0.713 & 0.725 & 0.885 & 0.687 & 0.747\\
     & M3DM~\cite{wang2023multimodal}& 0.941 & 0.651 & 0.965 & 0.969 & 0.905 & 0.760 & 0.880 & 0.974 & 0.926 & 0.765 & 0.874 \\
     & ShapeGuided~\cite{pmlr-v202-chu23b} & \textcolor{red}{0.983} & 0.682 & 0.978 & \textcolor{red}{0.998} & \textcolor{red}{0.960} & 0.737 & \textcolor{red}{0.993} & \textcolor{red}{0.979} & 0.966 & 0.871 & 0.916\\
     & 3DSR~\cite{CD} & 0.945 & \textcolor{red}{0.835} & 0.969 & 0.857 & 0.955 & 0.880 & 0.963 & 0.934 &  0.998 & \textcolor{red}{0.888} & 0.922\\
     & DAS3D & 0.927 & 0.750 & \textcolor{red}{0.986} & 0.997 & 0.954 & \textcolor{red}{0.905} & 0.958 & 0.902 & \textcolor{red}{1.000} & 0.854 & \textcolor{red}{0.923}\\[2pt]
     \hline
    \multirow{6}{*}{RGB+3D} & BTF\cite{horwitz2022feature} & 0.938 &0.765 &0.972 &0.888 &0.960 &0.664 &0.904 &0.929 &0.982 &0.726 &0.873\\
     & AST\cite{RudWeh2023} & 0.983 &0.873 &0.976 &0.971 &0.932 &0.885 &0.974 &0.981 &\textcolor{red}{1.000} &0.797& 0.937\\
     & EasyNet\cite{easynet} & 0.991 &\textcolor{red}{0.998} &0.918 &0.968 &0.945 &0.945 &0.905 &0.807 &0.994 &0.793 &0.926\\
     & M3DM~\cite{wang2023multimodal}& 0.994 &0.909 &0.972 &0.976 &0.960 &0.942 &0.973 &0.899 &0.972 &0.850 &0.945 \\
     & ShapeGuided~\cite{pmlr-v202-chu23b} & 0.986 &0.894 &0.983 &0.991 &0.976 &0.857 &\textcolor{red}{0.990} &0.965 &0.960 &0.869 &0.947\\
     & 3DSR~\cite{CD} & 0.981 & 0.867 & 0.996 & 0.981 & \textcolor{red}{1.000} & 0.994 & 0.986 & \textcolor{red}{0.978} & \textcolor{red}{1.000} &  \textcolor{red}{0.995} & 0.978\\
     & DAS3D & \textcolor{red}{0.997} &0.973 &\textcolor{red}{0.999} &\textcolor{red}{0.992} & 0.970 &\textcolor{red}{0.995} &0.962 &0.954 &0.998 &0.977 & \textcolor{red}{0.982}\\
    \bottomrule
\end{tabular}
\vspace{-2.5mm}
\end{table}

\begin{table}[!ht]
\fontsize{6}{8}\selectfont
\setlength{\tabcolsep}{0.6mm}
\renewcommand{\arraystretch}{1.1}
    \centering
    \caption{AUPRO score for anomaly segmentation of MVTec 3D-AD. The best is in red.}
    \label{tab:aupro-mvtec}
    \begin{tabular}{cl|cccccccccc|c}
    \toprule
    \multicolumn{2}{c|}{Method} & Bagel & Cable & Carrot & Cookie &Dowel & Foam & Peach & Potato & Rope & Tire & Mean\\
    \hline
    \multirow{4}{*}{RGB only} 
     & EasyNet~\cite{easynet} & 0.751 & 0.825 & 0.916 & 0.599 & 0.698 & 0.699 & 0.917 & 0.827 & 0.887 & 0.636 & 0.776\\
     & M3DM~\cite{wang2023multimodal} & \textcolor{red}{0.952} & \textcolor{red}{0.972} & 0.973 & 0.891 & 0.932 & 0.843 & \textcolor{red}{0.970} & \textcolor{red}{0.956} & \textcolor{red}{0.968} & 0.966 & 0.942\\
     & ShapeGuided~\cite{pmlr-v202-chu23b} & 0.946 & \textcolor{red}{0.972} & 0.960 & \textcolor{red}{0.914} & 0.958 & 0.776 & 0.937 & 0.949 & 0.956 & 0.957 & 0.933\\
     & 3DSR~\cite{CD} & 0.923 & 0.970 & \textcolor{red}{0.979} & 0.859 & \textcolor{red}{0.979} & \textcolor{red}{0.894} & 0.943 & 0.951 & 0.964 & \textcolor{red}{0.980} & \textcolor{red}{0.944}\\

     & DAS3D & 0.909 & 0.884 & 0.964 & 0.784 & 0.915 & 0.837 & 0.921 & 0.925 & 0.949 & 0.967 & 0.906\\
     \hline
    \multirow{4}{*}{3D only} 
    & EasyNet~\cite{easynet} & 0.160 & 0.030 & 0.680 & 0.759 & 0.758 & 0.069 & 0.225 & 0.734 & 0.797 & 0.509 & 0.472\\
     & M3DM~\cite{wang2023multimodal} & 0.943 & 0.818 & 0.977 & 0.882 & 0.881 & 0.743 & 0.958 & 0.974 & 0.950 & 0.929 & 0.906\\
     & ShapeGuided~\cite{pmlr-v202-chu23b} & \textcolor{red}{0.974} & 0.871 & 0.981 & 0.924 & 0.898 & 0.773 & 0.978 & \textcolor{red}{0.983} & 0.955 & 0.969 & 0.931\\
     & 3DSR~\cite{CD} &  0.922 & 0.872 & \textcolor{red}{0.984} & 0.859 & \textcolor{red}{0.940} & 0.714 & 0.970 & 0.978 & \textcolor{red}{0.977} & 0.858 & 0.907\\
     & DAS3D & 0.959 & \textcolor{red}{0.923} & 0.981 & \textcolor{red}{0.970} & 0.935 & \textcolor{red}{0.831} & \textcolor{red}{0.979} & 0.982 & 0.974 & \textcolor{red}{0.981} & \textcolor{red}{0.952}\\
     \hline
    \multirow{4}{*}{RGB+3D} 
     & EasyNet~\cite{easynet} & 0.839 & 0.864 & 0.951 & 0.618 & 0.828 & 0.836 & 0.942 & 0.889 & 0.911 & 0.528 & 0.821\\
     & M3DM~\cite{wang2023multimodal} & 0.970 & 0.971 & 0.979 & 0.950 & 0.941 & 0.932 & 0.977 & 0.971 & 0.971 & 0.975 & 0.964\\
     & ShapeGuided~\cite{pmlr-v202-chu23b} & \textcolor{red}{0.981} & \textcolor{red}{0.973} & 0.982 & \textcolor{red}{0.971} & 0.962 & 0.978 & \textcolor{red}{0.981} & \textcolor{red}{0.983} & 0.974 & 0.975 & \textcolor{red}{0.976}\\
     & 3DSR~\cite{CD} & 0.964 & 0.966 & 0.981 & 0.942 & \textcolor{red}{0.980} & 0.973 & \textcolor{red}{0.981} & 0.977 & \textcolor{red}{0.979} & 0.979 & 0.972\\

     & DAS3D & \textcolor{red}{0.981} & 0.950 & \textcolor{red}{0.983} & 0.968 & 0.961 & \textcolor{red}{0.983} & \textcolor{red}{0.981} & 0.980 & 0.978 & \textcolor{red}{0.983} & 0.975\\
    \bottomrule
\end{tabular}
\vspace{-15.0pt}
\end{table}

\section{Experiments}
\label{sec:Exp}

\subsection{Experimental Setup}

\noindent \textbf{Datasets.} We evaluate our method on MVTec 3D-AD~\cite{Bergmann_2022} and Eyescandies~\cite{bonfiglioli2022eyecandies}. The MVTec 3D-AD dataset is the first 3D industrial anomaly detection dataset, which consists of 10 categories, a total of 2656 training samples, and 1137 testing samples. The Eyecandies dataset is a novel synthetic dataset for unsupervised 3D anomaly detection and localization, which also consists of 10 categories, a total of 50,000 training samples, and 2,500 testing samples. For a more detailed introduction to these two datasets, please refer to the supplementary materials.

\noindent \textbf{Data Preprocess.} We remove the background information through 3D data. Following previous methods~\cite{wang2023multimodal,pmlr-v202-chu23b}, we estimate the background plane with RANSAC~\cite{fischler1981random} and any point within 0.005 distance is removed. The depth of the removed points is set as the maximum depth among the remaining points, and their corresponding pixel values in the RGB image are set to 0. we resize both the depth and RGB images to 256$\times$256 and normalize them into (0,1).

\noindent \textbf{Metrics.} 
We evaluate the image-level anomaly detection performance with the area under the receiver operator curve (I-AUROC), and higher I-AUROC mean better image-level anomaly detection performance. Similar to I-AUROC, the receiver operator curve of pixel level predictions can be used to calculate P-AUROC for evaluating the segmentation performance. We also use the per-region overlap (AUPRO) metric to evaluate segmentation performance, which is defined as the average relative overlap of the binary prediction with each connected component of the ground truth.

\begin{table}[t]
\fontsize{6}{8}\selectfont
\setlength{\tabcolsep}{0.6mm}
\renewcommand{\arraystretch}{1.1}
    \centering
    \caption{I-AUROC score for anomaly detection of Eyescandies. The best is in red.}
    \label{tab:i-auroc-candy}
    \begin{tabular}{cl|cccccccccc|c}
    \toprule
    \multicolumn{2}{c|}{\multirow{2}{*}{Method}} & Candy & Choc. & Choc. & \multirow{2}{*}{Conf.} & Gummy & Haze. & Lico. & \multirow{2}{*}{Loll.} & \multirow{2}{*}{Mars.} & Pepp. & Mean\\
    & & Cane & Cook. & Pral. & & Bear & Truf. & Sand. & & & Candy\\
    \hline
    \multirow{6}{*}{RGB only} 
    & STEPM~\cite{DBLP:conf/bmvc/WangHD021} & 0.551 & 0.654 & 0.576 & 0.784 & 0.737 & 0.790 & 0.778 & 0.620 & 0.840 & 0.749 & 0.708\\
    & PaDiM~\cite{defard2021padim} & 0.531 & 0.816 & 0.821 & 0.856 & 0.826 & 0.727 & 0.784 & 0.665 & 0.987 & 0.924 & 0.794\\
    & AutoEncoder~\cite{bonfiglioli2022eyecandies} & 0.527 & 0.848 & 0.772 & 0.734 & 0.590 & 0.508 & 0.693 & 0.760 & 0.851 & 0.730 & 0.701\\
    & M3DM~\cite{wang2023multimodal} & 0.648 & 0.949 & 0.941 & \textcolor{red}{1.000} & \textcolor{red}{0.878} & 0.632 & \textcolor{red}{0.933} & 0.811 & \textcolor{red}{0.998} & \textcolor{red}{1.000} & 0.879\\
     & EasyNet~\cite{easynet} & \textcolor{red}{0.723} & 0.925 & 0.849 & 0.966 & 0.705 & \textcolor{red}{0.815} & 0.806 & 0.851 & 0.975 & 0.960 & 0.858\\
     & 3DSR~\cite{CD} & 0.706 & \textcolor{red}{0.965} & \textcolor{red}{0.950} & 0.966 & 0.870 & 0.790 & 0.885 & \textcolor{red}{0.857} & \textcolor{red}{0.998} & 0.992 & \textcolor{red}{0.898} \\
     & DAS3D & 0.690 & 0.967 & 0.873 & 0.971 & 0.724 & 0.674 & 0.632 & 0.631 & 0.669 & 0.975 & 0.781\\
     \hline
    \multirow{5}{*}{3D only} 
    & SIFT~\cite{DBLP:journals/corr/abs-2203-05550} & 0.589 & 0.582 & 0.683 & 0.885 & 0.663 & 0.480 & 0.778 & 0.702 & 0.746 & 0.790 & 0.690\\
    & FPFH~\cite{DBLP:journals/corr/abs-2203-05550} & 0.670 & 0.710 & 0.805 & 0.806 & 0.748 & 0.515 & 0.794 & 0.757 & 0.765 & 0.757 & 0.733\\
    & M3DM~\cite{wang2023multimodal} & 0.482 & 0.589 & \textcolor{red}{0.805} & 0.845 & 0.780 & 0.538 & 0.766 & 0.827 & 0.800 & 0.822 & 0.725\\
     & EasyNet~\cite{easynet} & 0.629 & 0.716 & 0.768 & 0.731 & 0.660 & 0.710 & 0.712 & 0.711 & 0.688 & 0.731 & 0.706\\
     & 3DSR~\cite{CD} & 0.600 & 0.768 & 0.742 & 0.770 & 0.761 & 0.749 & 0.811 & \textcolor{red}{0.831} & 0.811 & 0.917 & 0.776\\
     & DAS3D & \textcolor{red}{0.718} & \textcolor{red}{0.808} & 0.768 & \textcolor{red}{0.906} & \textcolor{red}{0.816} & \textcolor{red}{0.763} & \textcolor{red}{0.858} & 0.812 & \textcolor{red}{0.862} & \textcolor{red}{0.933} & \textcolor{red}{0.824}\\
     \hline
    \multirow{5}{*}{RGB+3D} & AutoEncoder~\cite{bonfiglioli2022eyecandies} & 0.529 & 0.861 & 0.739 & 0.752 & 0.594 & 0.498 & 0.679 & 0.651 & 0.838 & 0.750 & 0.689\\
    & FPFH~\cite{DBLP:journals/corr/abs-2203-05550} & 0.606 & 0.904 & 0.792 & 0.939 & 0.720 & 0.563 & 0.867 & 0.860 & 0.992 & 0.842 & 0.809\\
    & M3DM~\cite{wang2023multimodal} & 0.624 & 0.958 & \textcolor{red}{0.958} & \textcolor{red}{1.000} & \textcolor{red}{0.886} & 0.758 & 0.949 & 0.836 & \textcolor{red}{1.000} & \textcolor{red}{1.000} & 0.897\\
     & EasyNet~\cite{easynet} & 0.737 & 0.934 & 0.866 & 0.966 & 0.717 & 0.822 & 0.847 & 0.863 & 0.977 & 0.960 & 0.869\\
     & 3DSR~\cite{CD} & 0.651 & \textcolor{red}{0.998} & 0.904 & 0.978 & 0.875 & 0.861 & \textcolor{red}{0.965} & \textcolor{red}{0.899} & 0.990 & 0.971 & 0.909
\\
     & DAS3D & \textcolor{red}{0.780} & 0.972 & 0.900 & 0.970 & 0.881 & \textcolor{red}{0.884} & 0.925 & 0.860 & 0.990 & 0.988 & \textcolor{red}{0.915}\\
    \bottomrule
\end{tabular}
\vspace{-5.0pt}
\end{table}

\begin{table}[t]
\fontsize{6}{8}\selectfont
\setlength{\tabcolsep}{0.6mm}
\renewcommand{\arraystretch}{1.1}
    \centering
    \caption{AUPRO score for anomaly segmentation of Eyescandies. The best is in red.}
    \label{tab:aupro-candy}
    \begin{tabular}{cl|cccccccccc|c}
    \toprule
    \multicolumn{2}{c|}{\multirow{2}{*}{Method}} & Candy & Choc. & Choc. & \multirow{2}{*}{Conf.} & Gummy & Haze. & Lico. & \multirow{2}{*}{Loll.} & \multirow{2}{*}{Mars.} & Pepp. & \multirow{2}{*}{Mean}\\
    & & Cane & Cook. & Pral. & & Bear & Truf. & Sand. & & & Candy\\
    \hline
    \multirow{3}{*}{RGB only} 
     & EasyNet~\cite{easynet} & 0.899 & 0.796 & 0.832 & 0.939 & 0.820 & 0.643 & 0.914 & 0.865 & 0.947 & 0.933 & 0.859\\
     & M3DM~\cite{wang2023multimodal} & 0.867 & \textcolor{red}{0.904} & 0.805 & \textcolor{red}{0.982} & \textcolor{red}{0.871} & 0.662 & 0.882 & \textcolor{red}{0.895} & \textcolor{red}{0.970} & 0.962 & 0.880\\
     & DAS3D & \textcolor{red}{0.910} & 0.866 & \textcolor{red}{0.906} & 0.970 & 0.845 & \textcolor{red}{0.741} & \textcolor{red}{0.917} & 0.887 & 0.950 & \textcolor{red}{0.975} & \textcolor{red}{0.897}\\
     \hline
    \multirow{3}{*}{3D only} 
    & EasyNet~\cite{easynet} & 0.489 & 0.368 & 0.488 & 0.614 & 0.557 & 0.362 & 0.515 & 0.740 & 0.627 & 0.601 & 0.536\\
     & M3DM~\cite{wang2023multimodal} & \textcolor{red}{0.911} & \textcolor{red}{0.645} & 0.581 & 0.748 & \textcolor{red}{0.748} & 0.484 & \textcolor{red}{0.608} & \textcolor{red}{0.904} & 0.646 & \textcolor{red}{0.750} & 0.702\\
     & DAS3D & 0.712 & 0.530 & \textcolor{red}{0.658} & \textcolor{red}{0.792} & 0.688 & \textcolor{red}{0.622} & 0.603 & 0.851 & \textcolor{red}{0.721} & 0.736 & \textcolor{red}{0.721}\\
     \hline
    \multirow{3}{*}{RGB+3D} 
     & EasyNet~\cite{easynet} & 0.895 & 0.796 & 0.710 & 0.862 & 0.820 & 0.465 & 0.827 & 0.701 & 0.956 & 0.897 & 0.793\\
     & M3DM~\cite{wang2023multimodal} & 0.906 & \textcolor{red}{0.923} & 0.803 & \textcolor{red}{0.983} & 0.855 & 0.688 & \textcolor{red}{0.880} & 0.906 & 0.966 & 0.955 & 0.882\\
     & DAS3D & \textcolor{red}{0.932} & 0.901 & \textcolor{red}{0.904} & 0.982 & \textcolor{red}{0.939} & \textcolor{red}{0.846} & 0.866 & \textcolor{red}{0.942} & \textcolor{red}{0.976} & \textcolor{red}{0.979} & \textcolor{red}{0.927}\\
    \bottomrule
\end{tabular}
\vspace{-10.0pt}
\end{table}

\subsection{Main Results}

\noindent \textbf{Quantitative Results}. 
Anomaly detection results on the MVTec 3D-AD are presented in Table~\ref{tab:i-auroc-mvtec} and Table~\ref{tab:aupro-mvtec}.
In comparison with the SOTA method~\cite{pmlr-v202-chu23b}, our approach obtain an absolute AUROC gain of $3.4\%$ for the image-level detection. Besides, our method also achieve competitive segmentation performance, which is 0.3\% lower than ShapeGuided. More comparison results on AUPRO are provided in supplementary materials.
Table~\ref{tab:i-auroc-candy} shows the detection results on the Eyescandies dataset. Our method still achieves the best detection performance and outperforms M3DM~\cite{wang2023multimodal} by 1.8\%. More quantitative segmentation results can be found in supplementary materials. On both datasets, our approach achieves new state-of-the-art performance on detection results, and surpasses other methods by a large margin. It is worth noting that our method achieves better detection and segmentation performance with RGB and 3D information than using only RGB or 3D, which verifies that our method fully exploits the information from both modalities to achieve enhanced detection.

\begin{figure}[t]
    \scriptsize
    \centering
    \begin{minipage}[t]{\linewidth}
    
    \begin{minipage}[t]{0.158\linewidth}
        \centering
        \centerline{\includegraphics[width=1\linewidth]{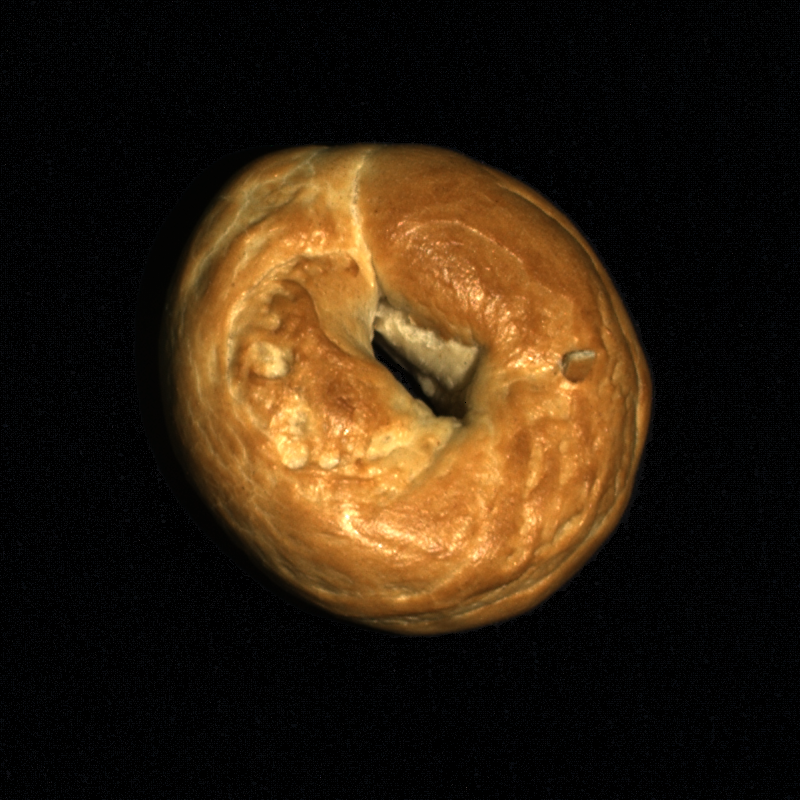}}
    \end{minipage}
    \begin{minipage}[t]{0.158\linewidth}
        \centering
        \centerline{\includegraphics[width=1\linewidth]{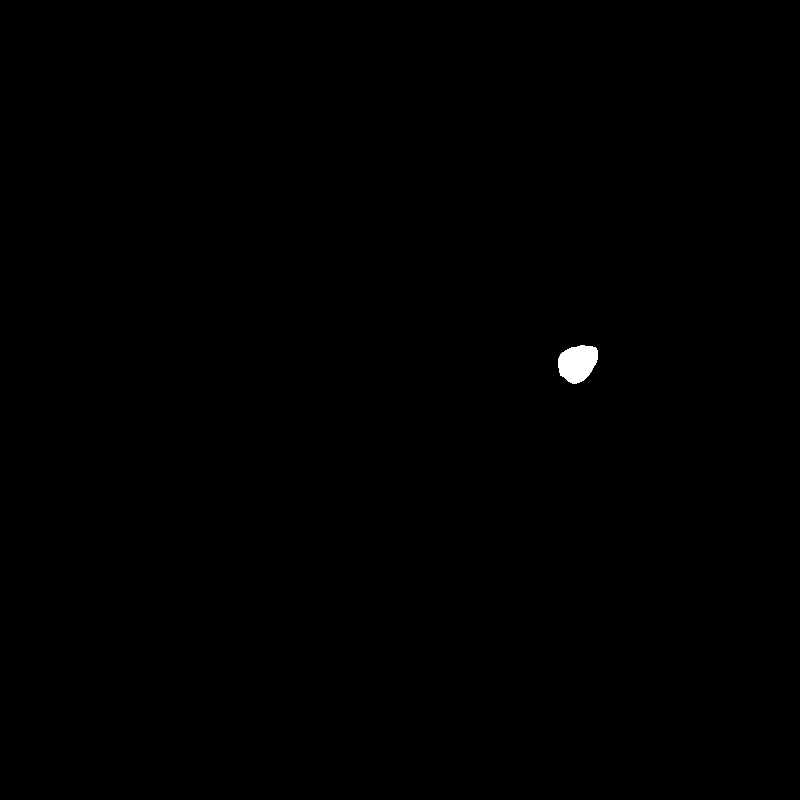}}
    \end{minipage}
    \begin{minipage}[t]{0.158\linewidth}
        \centering
        \centerline{\includegraphics[width=1\linewidth]{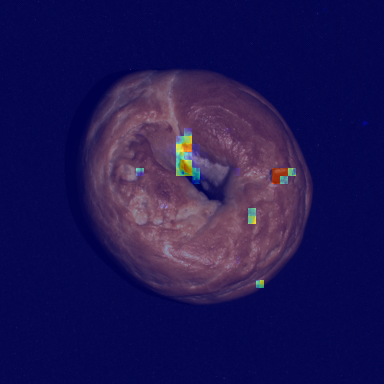}}
    \end{minipage}
    \begin{minipage}[t]{0.158\linewidth}
        \centering
        \centerline{\includegraphics[width=1\linewidth]{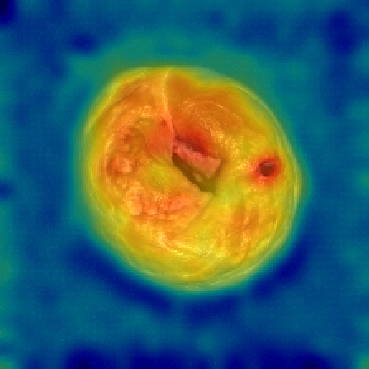}}
    \end{minipage}
    \begin{minipage}[t]{0.158\linewidth}
        \centering
        \centerline{\includegraphics[width=1\linewidth]{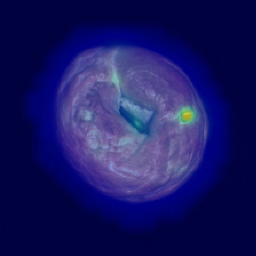}}
    \end{minipage}
    \begin{minipage}[t]{0.158\linewidth}
        \centering
        \centerline{\includegraphics[width=1\linewidth]{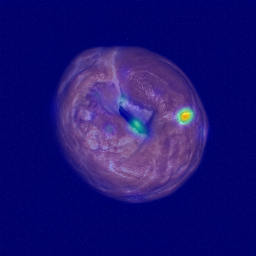}}
    \end{minipage}
        
        \begin{minipage}[t]{0.158\linewidth}
            \centering
            \centerline{\includegraphics[width=1\linewidth]{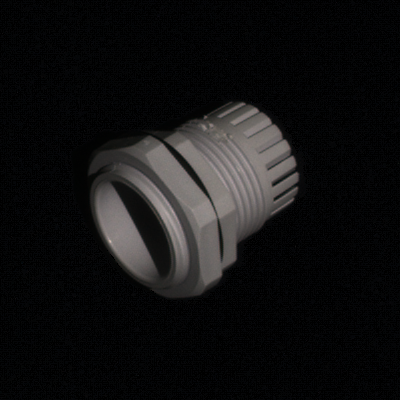}}
        \end{minipage}
        \begin{minipage}[t]{0.158\linewidth}
            \centering
            \centerline{\includegraphics[width=1\linewidth]{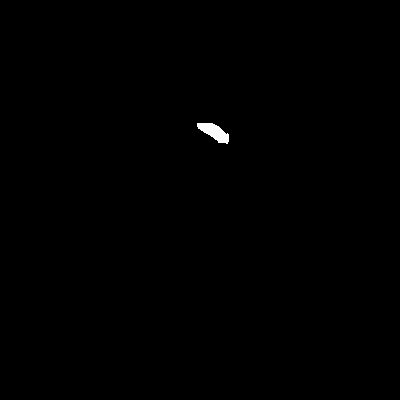}}
        \end{minipage}
        \begin{minipage}[t]{0.158\linewidth}
            \centering
            \centerline{\includegraphics[width=1\linewidth]{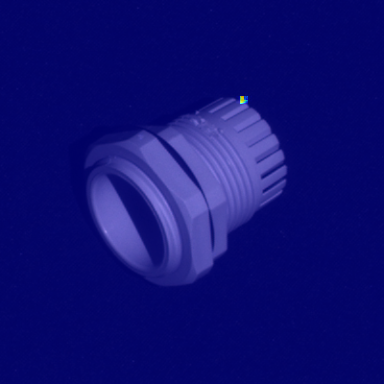}}
        \end{minipage}
        \begin{minipage}[t]{0.158\linewidth}
            \centering
            \centerline{\includegraphics[width=1\linewidth]{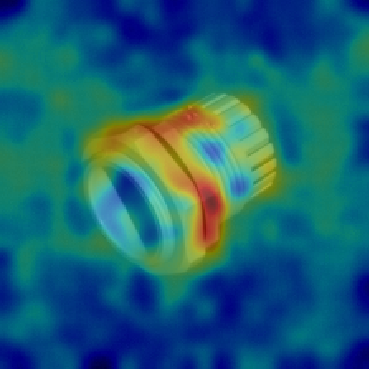}}
        \end{minipage}
        \begin{minipage}[t]{0.158\linewidth}
            \centering
            \centerline{\includegraphics[width=1\linewidth]{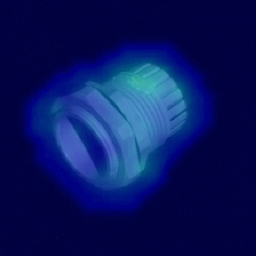}}
        \end{minipage}
        \begin{minipage}[t]{0.158\linewidth}
            \centering
            \centerline{\includegraphics[width=1\linewidth]{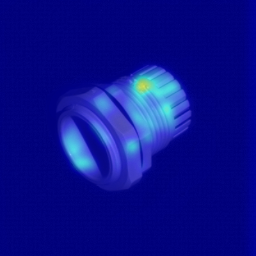}}
        \end{minipage}

    
        \begin{minipage}[t]{0.158\linewidth}
            \centering
            \centerline{\includegraphics[width=1\linewidth]{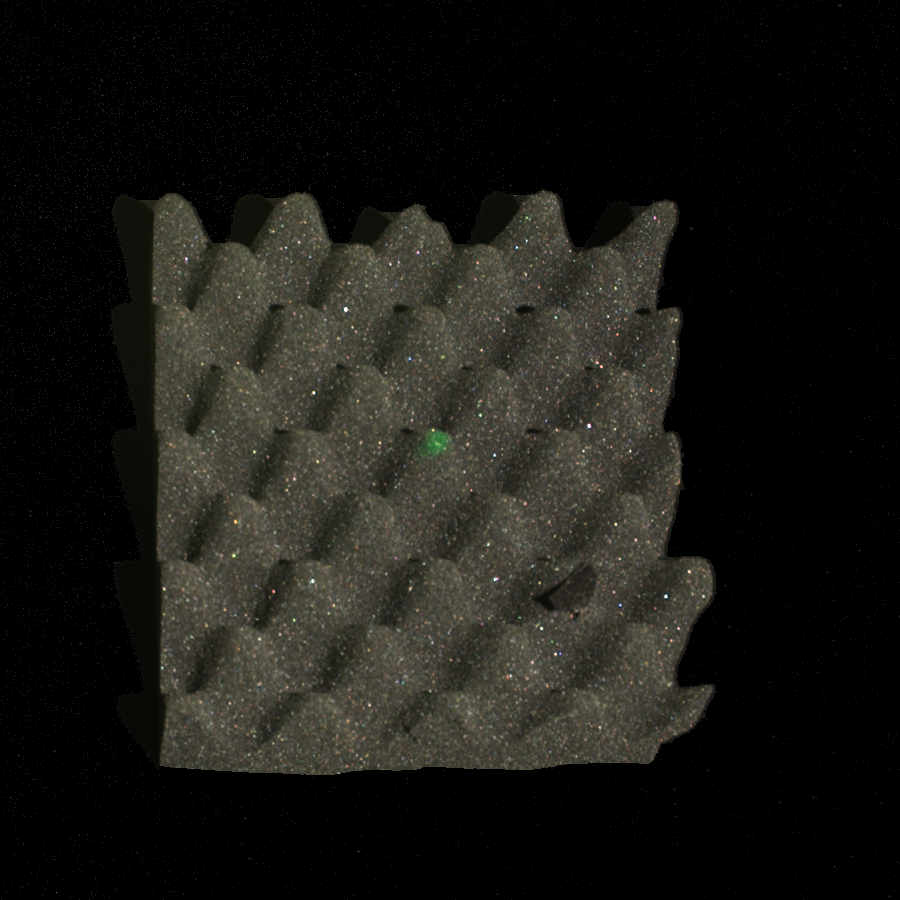}}
        \end{minipage}
        \begin{minipage}[t]{0.158\linewidth}
            \centering
            \centerline{\includegraphics[width=1\linewidth]{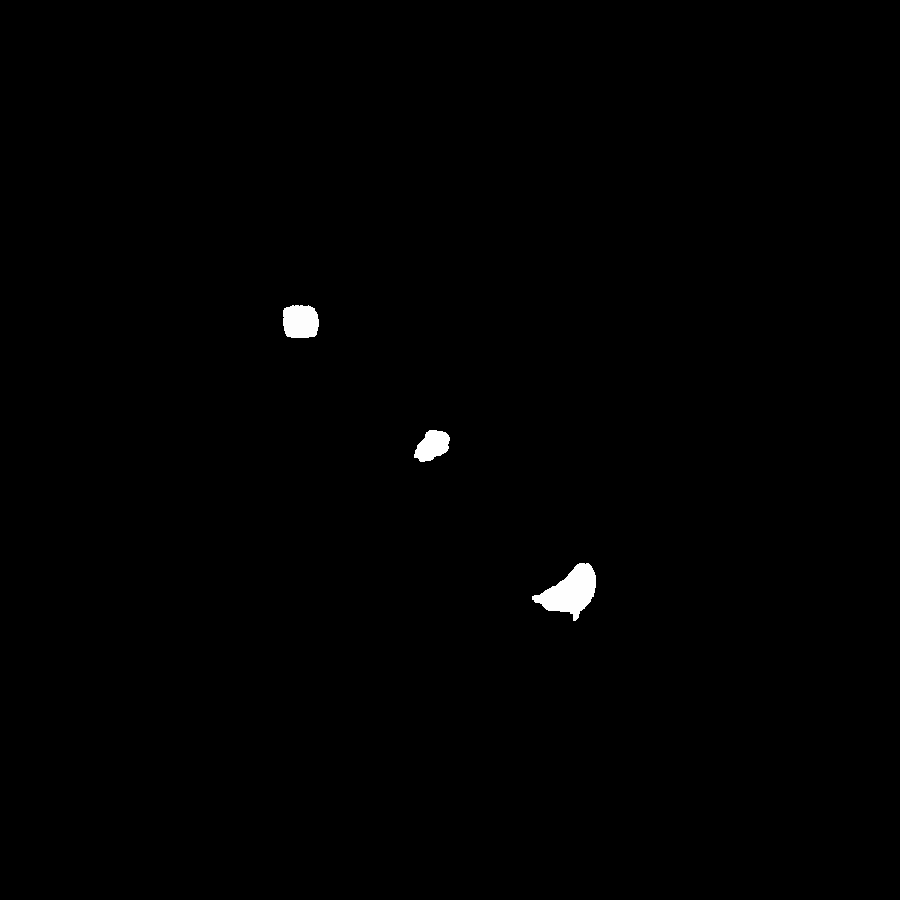}}
        \end{minipage}
        \begin{minipage}[t]{0.158\linewidth}
            \centering
            \centerline{\includegraphics[width=1\linewidth]{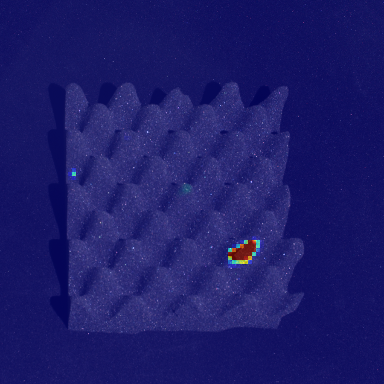}}
        \end{minipage}
        \begin{minipage}[t]{0.158\linewidth}
            \centering
            \centerline{\includegraphics[width=1\linewidth]{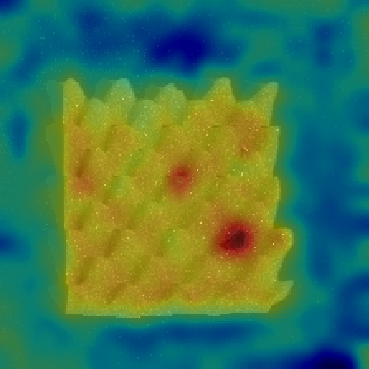}}
        \end{minipage}
        \begin{minipage}[t]{0.158\linewidth}
            \centering
            \centerline{\includegraphics[width=1\linewidth]{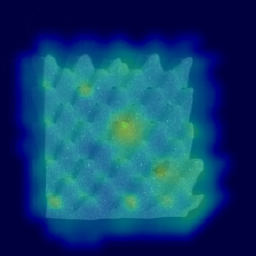}}
        \end{minipage}
        \begin{minipage}[t]{0.158\linewidth}
            \centering
            \centerline{\includegraphics[width=1\linewidth]{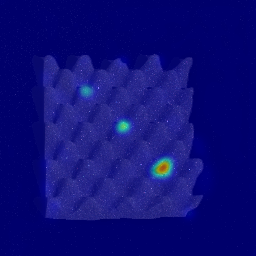}}
        \end{minipage}

        \begin{minipage}[t]{0.158\linewidth}
            \centering
            \centerline{\includegraphics[width=1\linewidth]{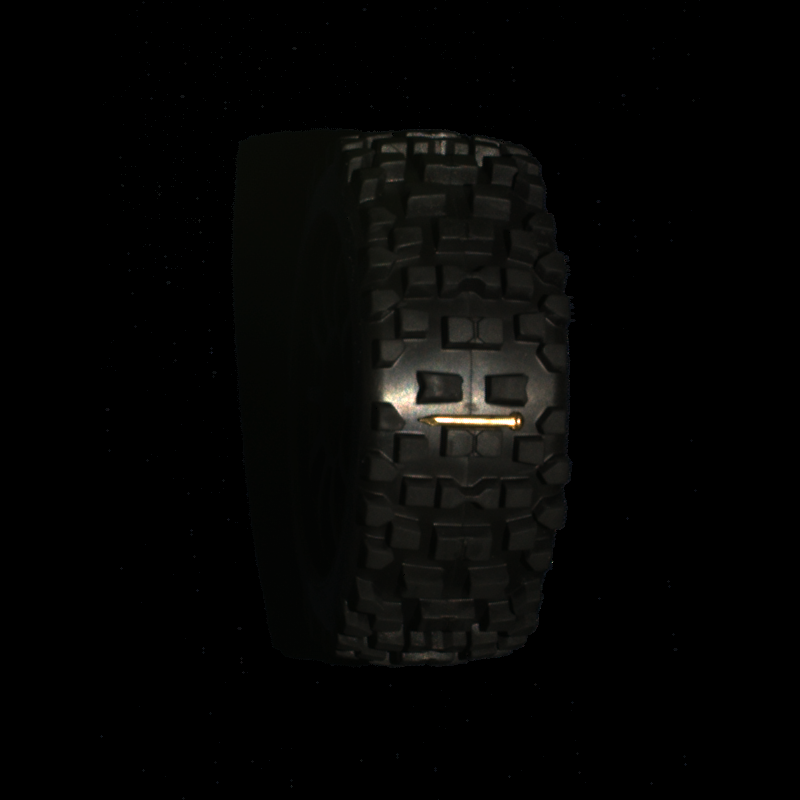}}
            \centerline{RGB}
        \end{minipage}
        \begin{minipage}[t]{0.158\linewidth}
            \centering
            \centerline{\includegraphics[width=1\linewidth]{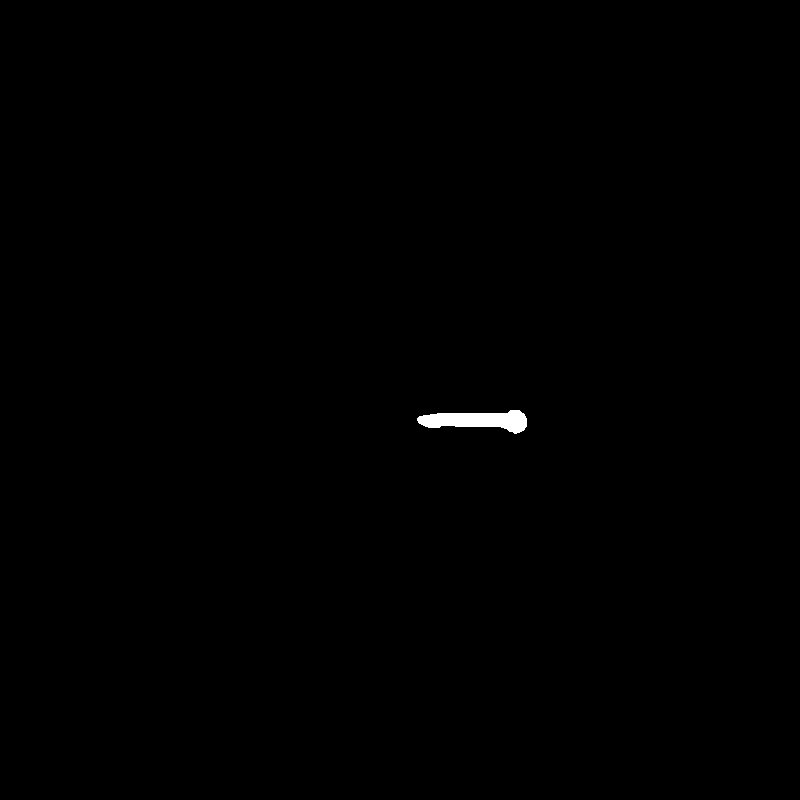}}
            \centerline{GT}
        \end{minipage}
        \begin{minipage}[t]{0.158\linewidth}
            \centering
            \centerline{\includegraphics[width=1\linewidth]{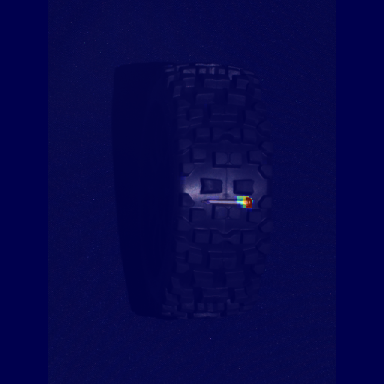}}
            \centerline{3DSR}
        \end{minipage}
        \begin{minipage}[t]{0.158\linewidth}
            \centering
            \centerline{\includegraphics[width=1\linewidth]{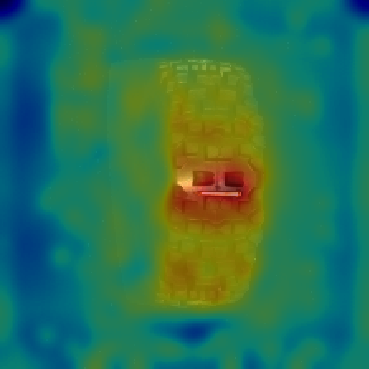}}
            \centerline{M3DM}
        \end{minipage}
        \begin{minipage}[t]{0.158\linewidth}
            \centering
            \centerline{\includegraphics[width=1\linewidth]{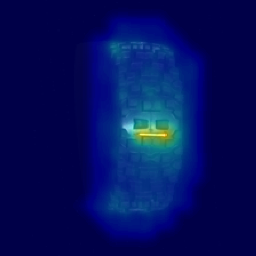}}
            \centerline{ShapeGuided}
        \end{minipage}
        \begin{minipage}[t]{0.158\linewidth}
            \centering
            \centerline{\includegraphics[width=1\linewidth]{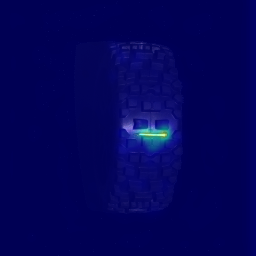}}
            \centerline{DAS3D}
        \end{minipage}
    \end{minipage}
    \caption{Anomaly detection on different categories. From top to bottom: bagel, cable gland, foam and tire. From left to right: RGB image, ground truth, anomaly maps of 3DSR~\cite{CD}, M3DM~\cite{wang2023multimodal}, ShapeGuided~\cite{pmlr-v202-chu23b} and DAS3D. The red color corresponds to high anomaly score, whereas the blue represents low anomaly score. Best view in color.}
    \label{fig:example}
    \vspace{-5mm}
\end{figure}

\noindent \textbf{Visualization Results}.
Figure~\ref{fig:example} illustrates four challenging cases where existing methods fail to accurately localize all defects. These instances involve extremely small defects or cases with multiple defects. Since small defects are usually inconspicuous, the unevenness on the object surface can easily mislead anomaly map predictions. For instance, in the bagel case (first row), due to the defect's small size, M3DM and 3DSR pay more attention to the uneven region in the upper left part. However, our method accurately predicts the anomaly map thanks to our diverse synthetic anomaly samples. Color defects are typically detected only from RGB information, while cut defects are difficult to be detected from the RGB images with complex textures. In the foam case (third row), the defects contain a color defect (center), a cut defect (upper left) and a contamination defect (lower right). ShapeGuided cannot provide useful predictions due to the complex shape of foam. M3DM identifies the color and contamination defects but misses the cut one. 3DSR only identifies the contamination defects and misses the other two. Benefiting from the augmentation dropout mechanism, our discriminator effectively retain useful information from different modalities, leading to better prediction.

\noindent \textbf{Complexity Analysis}.
Inference speed and memory usage are important in industrial applications. Our method merely contains three UNet-like networks and the inference can be completed in a single forward pass. As demonstrated in Table~\ref{tab:ComCom}, our method achieves faster inference speed, being $60\times$ faster than M3DM~\cite{wang2023multimodal} and $140\times$ faster than ShapeGuided~\cite{pmlr-v202-chu23b}, respectively, while obtains better detection precision and competitive segmentation results. Although the inference time and memory usage of 3DSR~\cite{CD} is competitive to our method, their detection and segmentation performance significantly underperforms in comparison. 

\begin{table}[t]
\begin{minipage}{0.5\linewidth}
\fontsize{6}{8}\selectfont
\setlength{\tabcolsep}{0.6mm}
\renewcommand{\arraystretch}{1.1}
\centering
    \caption{The comparison of different methods in terms of inference time (s), memory usage (MB) and performance (I-AUROC/AUPRO) on MVTec 3D-AD.}
    \label{tab:ComCom}
    \begin{tabular}{l|c|c|c}
    \toprule
        Method & Inf. time & Memory & Performance\\
        \hline
        3DSR~\cite{CD} & 0.049 & 3210 & 0.978/0.972\\
        M3DM~\cite{wang2023multimodal} & 2.495 & 11368 & 0.945/0.964\\
        ShapeGuided~\cite{pmlr-v202-chu23b} & 5.782 & 6485 & 0.947/\textcolor{red}{0.976}\\
        \hline
        DAS3D & 0.041 & 3936 & \textcolor{red}{0.982}/0.975\\
    \bottomrule
    \end{tabular}
\end{minipage}\hspace{2mm}
\begin{minipage}{0.5\linewidth}
\fontsize{6}{8}\selectfont
\setlength{\tabcolsep}{0.6mm}
\renewcommand{\arraystretch}{1.1}
\centering
\caption{Ablation study on our synthesis method and augmentation dropout mechanism. Combining two modules contributes to the best performance of DAS3D.}
    \label{tab:ablation}
    \begin{tabular}{cc|cc}
    \toprule
    Skew Filter & Dropout & I-AUROC & P-AUROC\\
    \hline
    & & 0.922 & 0.973\\
    \checkmark & & 0.959 & 0.977\\
    & \checkmark & 0.941 & 0.975\\
    \checkmark & \checkmark & \textcolor{red}{0.982} & \textcolor{red}{0.993}\\
    \bottomrule
    \end{tabular} 
\end{minipage}
\vspace{-16.0pt}
\end{table}


\subsection{Ablation Study}

We conduct an ablation study on MVTec 3D-AD dataset.
As Table~\ref{tab:ablation} shows, without using our synthesis method, we generate the anomaly depth through simply adding the augmentation mask $M_t$ (defined in Equation~\ref{eq:mt}) to the normal depth. Without the augmentation dropout mechanism, the dual-discriminator always learns from the augmented anomaly pairs, where both depth and RGB images are abnormal. As indicated in Table~\ref{tab:ablation}, both two modules contribute to improved performance and the combination of two modules achieves the best performance of DAS3D.

\section{Conclusion}

In this paper, we propose a novel 3D anomaly defects synthesis method, which is simple and effective. Our method draws inspiration from the synthesizing-based methods in RGB-based anomaly detection, which leverage the Perlin noise to generate useful defects regions. To effectively fuse the features from different modals, we design the augmentation dropout mechanism to enhance the generalizability of the discriminator. Our method surpasses the SOTA detection results on both MVTec-3D AD and Eyescandies dataset, while maintaining a rapid inference speed. We anticipate our work could provide valuable insights for future research in 3D anomaly detection tasks.

\clearpage  

%
%
\bibliographystyle{splncs04}
\bibliography{main}

\begin{thebibliography}{10}
\providecommand{\url}[1]{\texttt{#1}}
\providecommand{\urlprefix}{URL }
\providecommand{\doi}[1]{https://doi.org/#1}

\bibitem{skewnormal}
Azzalini, A., Capitanio, A.: Statistical applications of the multivariate skew normal distribution. Journal of the Royal Statistical Society: Series B (Statistical Methodology)  \textbf{61}(3),  579--602 (1999)

\bibitem{bergmann2022beyond}
Bergmann, P., Batzner, K., Fauser, M., Sattlegger, D., Steger, C.: Beyond dents and scratches: Logical constraints in unsupervised anomaly detection and localization. International Journal of Computer Vision  \textbf{130}(4),  947--969 (2022)

\bibitem{bergmann2019mvtec}
Bergmann, P., Fauser, M., Sattlegger, D., Steger, C.: Mvtec ad--a comprehensive real-world dataset for unsupervised anomaly detection. In: Proceedings of the IEEE/CVF conference on computer vision and pattern recognition. pp. 9592--9600 (2019)

\bibitem{Bergmann_2022}
Bergmann, P., Jin, X., Sattlegger, D., Steger, C.: The mvtec 3d-ad dataset for unsupervised 3d anomaly detection and localization. In: Proceedings of the 17th International Joint Conference on Computer Vision, Imaging and Computer Graphics Theory and Applications. SCITEPRESS - Science and Technology Publications (2022)

\bibitem{bergmann2018improving}
Bergmann, P., L{\"o}we, S., Fauser, M., Sattlegger, D., Steger, C.: Improving unsupervised defect segmentation by applying structural similarity to autoencoders. arXiv preprint arXiv:1807.02011  (2018)

\bibitem{bergmann2023anomaly}
Bergmann, P., Sattlegger, D.: Anomaly detection in 3d point clouds using deep geometric descriptors. In: Proceedings of the IEEE/CVF Winter Conference on Applications of Computer Vision. pp. 2613--2623 (2023)

\bibitem{bonfiglioli2022eyecandies}
Bonfiglioli, L., Toschi, M., Silvestri, D., Fioraio, N., De~Gregorio, D.: The eyecandies dataset for unsupervised multimodal anomaly detection and localization. In: Proceedings of the 16th Asian Conference on Computer Vision (ACCV2022 (2022), aCCV

\bibitem{easynet}
Chen, R., Xie, G., Liu, J., Wang, J., Luo, Z., Wang, J., Feng, Z.: Easynet: An easy network for 3d industrial anomaly detection. In: Proceedings of the 31st ACM International Conference on Multimedia. p. 7038–7046. MM '23 (2023)

\bibitem{pmlr-v202-chu23b}
Chu, Y.M., Liu, C., Hsieh, T.I., Chen, H.T., Liu, T.L.: Shape-guided dual-memory learning for 3d anomaly detection. In: Proceedings of the 40th International Conference on Machine Learning. pp. 6185--6194 (2023)

\bibitem{DTD}
Cimpoi, M., Maji, S., Kokkinos, I., Mohamed, S., Vedaldi, A.: Describing textures in the wild. In: 2014 IEEE Conference on Computer Vision and Pattern Recognition. pp. 3606--3613 (2014)

\bibitem{cohen2020sub}
Cohen, N., Hoshen, Y.: Sub-image anomaly detection with deep pyramid correspondences. arXiv preprint arXiv:2005.02357  (2020)

\bibitem{defard2021padim}
Defard, T., Setkov, A., Loesch, A., Audigier, R.: Padim: A patch distribution modeling framework for anomaly detection and localization. In: Del~Bimbo, A., Cucchiara, R., Sclaroff, S., Farinella, G.M., Mei, T., Bertini, M., Escalante, H.J., Vezzani, R. (eds.) Pattern Recognition. ICPR International Workshops and Challenges. pp. 475--489. Springer International Publishing, Cham (2021)

\bibitem{deng2022anomaly}
Deng, H., Li, X.: Anomaly detection via reverse distillation from one-class embedding. arXiv preprint arXiv:2201.10703  (2022)

\bibitem{deng2009imagenet}
Deng, J., Dong, W., Socher, R., Li, L.J., Li, K., Fei-Fei, L.: Imagenet: A large-scale hierarchical image database. In: 2009 IEEE Conference on Computer Vision and Pattern Recognition. pp. 248--255 (2009)

\bibitem{fischler1981random}
Fischler, M.A., Bolles, R.C.: Random sample consensus: a paradigm for model fitting with applications to image analysis and automated cartography. Communications of the ACM  \textbf{24}(6),  381--395 (1981)

\bibitem{Gu_2023_ICCV}
Gu, Z., Liu, L., Chen, X., Yi, R., Zhang, J., Wang, Y., Wang, C., Shu, A., Jiang, G., Ma, L.: Remembering normality: Memory-guided knowledge distillation for unsupervised anomaly detection. In: Proceedings of the IEEE/CVF International Conference on Computer Vision (ICCV). pp. 16401--16409 (October 2023)

\bibitem{gudovskiy2022cflow}
Gudovskiy, D., Ishizaka, S., Kozuka, K.: Cflow-ad: Real-time unsupervised anomaly detection with localization via conditional normalizing flows. In: Proceedings of the IEEE/CVF Winter Conference on Applications of Computer Vision (WACV). pp. 98--107 (January 2022)

\bibitem{Guo_2023_ICCV}
Guo, H., Ren, L., Fu, J., Wang, Y., Zhang, Z., Lan, C., Wang, H., Hou, X.: Template-guided hierarchical feature restoration for anomaly detection. In: Proceedings of the IEEE/CVF International Conference on Computer Vision (ICCV). pp. 6447--6458 (October 2023)

\bibitem{haselmann2019pixel}
Haselmann, M., Gruber, D.P.: Pixel-wise defect detection by cnns without manually labeled training data. Applied Artificial Intelligence  \textbf{33}(6),  548--566 (2019)

\bibitem{horwitz2022feature}
Horwitz, E., Hoshen, Y.: Back to the feature: Classical 3d features are (almost) all you need for 3d anomaly detection (2022)

\bibitem{DBLP:journals/corr/abs-2203-05550}
Horwitz, E., Hoshen, Y.: An empirical investigation of 3d anomaly detection and segmentation. CoRR  \textbf{abs/2203.05550} (2022)

\bibitem{huang2022registration}
Huang, C., Guan, H., Jiang, A., Zhang, Y., Spratling, M., Wang, Y.F.: Registration based few-shot anomaly detection. In: Avidan, S., Brostow, G., Ciss{\'e}, M., Farinella, G.M., Hassner, T. (eds.) Computer Vision -- ECCV 2022. pp. 303--319. Springer Nature Switzerland, Cham (2022)

\bibitem{kingma2013auto}
Kingma, D.P., Welling, M.: Auto-encoding variational bayes. arXiv preprint arXiv:1312.6114  (2013)

\bibitem{lecun1989generalization}
LeCun, Y., et~al.: Generalization and network design strategies. Connectionism in perspective  \textbf{19}(143-155), ~18 (1989)

\bibitem{li2021cutpaste}
Li, C.L., Sohn, K., Yoon, J., Pfister, T.: Cutpaste: Self-supervised learning for anomaly detection and localization. In: Proceedings of the IEEE/CVF Conference on Computer Vision and Pattern Recognition. pp. 9664--9674 (2021)

\bibitem{li2023scalable}
Li, W., Xu, X., Gu, Y., Zheng, B., Gao, S., Wu, Y.: Towards scalable 3d anomaly detection and localization: A benchmark via 3d anomaly synthesis and a self-supervised learning network (2023)

\bibitem{liang2022omni}
Liang, Y., Zhang, J., Zhao, S., Wu, R., Liu, Y., Pan, S.: Omni-frequency channel-selection representations for unsupervised anomaly detection. arXiv preprint arXiv:2203.00259  (2022)

\bibitem{lin2017focal}
Lin, T.Y., Goyal, P., Girshick, R., He, K., Doll{\'a}r, P.: Focal loss for dense object detection. In: Proceedings of the IEEE international conference on computer vision. pp. 2980--2988 (2017)

\bibitem{liu2023real3d}
Liu, J., Xie, G., Li, X., Wang, J., Liu, Y., Wang, C., Zheng, F., et~al.: Real3d-ad: A dataset of point cloud anomaly detection. In: Thirty-seventh Conference on Neural Information Processing Systems Datasets and Benchmarks Track (2023)

\bibitem{liu2022reconstruction}
Liu, T., Li, B., Zhao, Z., Du, X., Jiang, B., Geng, L.: Reconstruction from edge image combined with color and gradient difference for industrial surface anomaly detection (2022)

\bibitem{mei2018unsupervised}
Mei, S., Yang, H., Yin, Z.: An unsupervised-learning-based approach for automated defect inspection on textured surfaces. IEEE Transactions on Instrumentation and Measurement  \textbf{67}(6),  1266--1277 (2018)

\bibitem{nakazawa2019anomaly}
Nakazawa, T., Kulkarni, D.V.: Anomaly detection and segmentation for wafer defect patterns using deep convolutional encoder--decoder neural network architectures in semiconductor manufacturing. IEEE Transactions on Semiconductor Manufacturing  \textbf{32}(2),  250--256 (2019)

\bibitem{perlin}
Perlin, K.: An image synthesizer. In: Proceedings of the 12th Annual Conference on Computer Graphics and Interactive Techniques, {SIGGRAPH}. pp. 287--296. {ACM} (1985)

\bibitem{roth2022towards}
Roth, K., Pemula, L., Zepeda, J., Sch\"olkopf, B., Brox, T., Gehler, P.: Towards total recall in industrial anomaly detection. In: Proceedings of the IEEE/CVF Conference on Computer Vision and Pattern Recognition (CVPR). pp. 14318--14328 (June 2022)

\bibitem{rudolph2019structuring}
Rudolph, M., Wandt, B., Rosenhahn, B.: Structuring autoencoders. In: Proceedings of the IEEE/CVF International Conference on Computer Vision (ICCV) Workshops (Oct 2019)

\bibitem{rudolph2021same}
Rudolph, M., Wandt, B., Rosenhahn, B.: Same same but differnet: Semi-supervised defect detection with normalizing flows. In: Proceedings of the IEEE/CVF Winter Conference on Applications of Computer Vision (WACV). pp. 1907--1916 (January 2021)

\bibitem{RudWeh2023}
Rudolph, M., Wehrbein, T., Rosenhahn, B., Wandt, B.: Asymmetric student-teacher networks for industrial anomaly detection. In: Winter Conference on Applications of Computer Vision (WACV) (Jan 2023)

\bibitem{salehi2021multiresolution}
Salehi, M., Sadjadi, N., Baselizadeh, S., Rohban, M.H., Rabiee, H.R.: Multiresolution knowledge distillation for anomaly detection. In: Proceedings of the IEEE/CVF Conference on Computer Vision and Pattern Recognition. pp. 14902--14912 (2021)

\bibitem{schirrmeister2020understanding}
Schirrmeister, R., Zhou, Y., Ball, T., Zhang, D.: Understanding anomaly detection with deep invertible networks through hierarchies of distributions and features. In: Larochelle, H., Ranzato, M., Hadsell, R., Balcan, M., Lin, H. (eds.) Advances in Neural Information Processing Systems. vol.~33, pp. 21038--21049. Curran Associates, Inc. (2020), \url{https://proceedings.neurips.cc/paper/2020/file/f106b7f99d2cb30c3db1c3cc0fde9ccb-Paper.pdf}

\bibitem{schluter2022natural}
Schl{\"u}ter, H.M., Tan, J., Hou, B., Kainz, B.: Natural synthetic anomalies for self-supervised anomaly detection and localization. In: Computer Vision--ECCV 2022: 17th European Conference, Tel Aviv, Israel, October 23--27, 2022, Proceedings, Part XXXI. pp. 474--489. Springer (2022)

\bibitem{DBLP:conf/bmvc/WangHD021}
Wang, G., Han, S., Ding, E., Huang, D.: Student-teacher feature pyramid matching for anomaly detection. In: 32nd British Machine Vision Conference 2021, {BMVC} 2021, Online, November 22-25, 2021. p.~306. {BMVA} Press (2021)

\bibitem{wang2023multimodal}
Wang, Y., Peng, J., Zhang, J., Yi, R., Wang, Y., Wang, C.: Multimodal industrial anomaly detection via hybrid fusion. In: Proceedings of the IEEE/CVF Conference on Computer Vision and Pattern Recognition. pp. 8032--8041 (2023)

\bibitem{yan2021learning}
Yan, X., Zhang, H., Xu, X., Hu, X., Heng, P.A.: Learning semantic context from normal samples for unsupervised anomaly detection. In: Proceedings of the AAAI Conference on Artificial Intelligence. vol.~35, pp. 3110--3118 (2021)

\bibitem{yu2021fastflow}
Yu, J., Zheng, Y., Wang, X., Li, W., Wu, Y., Zhao, R., Wu, L.: Fastflow: Unsupervised anomaly detection and localization via 2d normalizing flows. arXiv preprint arXiv:2111.07677  (2021)

\bibitem{CD}
Z., V., K., M., S., D.: Cheating depth: Enhancing 3d surface anomaly detection via depth simulation. In: WACV (2024)

\bibitem{zavrtanik2021draem}
Zavrtanik, V., Kristan, M., Sko{\v{c}}aj, D.: Draem-a discriminatively trained reconstruction embedding for surface anomaly detection. In: Proceedings of the IEEE/CVF International Conference on Computer Vision. pp. 8330--8339 (2021)

\bibitem{zavrtanik2021reconstruction}
Zavrtanik, V., Kristan, M., Sko{\v{c}}aj, D.: Reconstruction by inpainting for visual anomaly detection. Pattern Recognition  \textbf{112},  107706 (2021)

\bibitem{zhang2023destseg}
Zhang, X., Li, S., Li, X., Huang, P., Shan, J., Chen, T.: Destseg: Segmentation guided denoising student-teacher for anomaly detection. In: Proceedings of the IEEE/CVF Conference on Computer Vision and Pattern Recognition. pp. 3914--3923 (2023)

\bibitem{zheng2022focus}
Zheng, Y., Wang, X., Deng, R., Bao, T., Zhao, R., Wu, L.: Focus your distribution: Coarse-to-fine non-contrastive learning for anomaly detection and localization. In: 2022 IEEE International Conference on Multimedia and Expo (ICME). pp.~1--6 (2022). \doi{10.1109/ICME52920.2022.9859925}

\bibitem{zong2018deep}
Zong, B., Song, Q., Min, M.R., Cheng, W., Lumezanu, C., Cho, D., Chen, H.: Deep autoencoding gaussian mixture model for unsupervised anomaly detection. In: International Conference on Learning Representations (2018), \url{https://openreview.net/forum?id=BJJLHbb0-}

\end{thebibliography}
\end{document}